\documentclass[10.5pt,reqno]{article}
\usepackage[margin=0.65in]{geometry} 
\usepackage[colorlinks,
            linkcolor=black,
            anchorcolor=black, 
            citecolor=black, 
            urlcolor=blue,
            ]{hyperref}     
\usepackage{indentfirst}
\usepackage{url}            
\usepackage{booktabs}       
\usepackage{amsfonts}       
\usepackage{nicefrac}       
\usepackage{microtype}      
\usepackage[reqno]{amsmath}
\usepackage{bbm}
\usepackage{graphicx}
\usepackage{subcaption}
\usepackage{wrapfig}
\usepackage{enumitem}
\usepackage{multirow}
\usepackage[font=small,labelfont=bf]{caption}
\usepackage{colortbl}
\usepackage[scaled]{helvet}
\usepackage{courier} 
\usepackage{lineno}
\usepackage{bbm}
\usepackage{color}
\usepackage{xcolor}
\usepackage{ulem}
\usepackage{nicematrix}
\newcommand{\ours}{\textsc{InstructCell}}

\definecolor{aliceblue}{RGB}{178, 217, 245}

\definecolor{babyblue}{RGB}{217, 239, 251}

\usepackage[fixed]{fontawesome5}

\usepackage{textcomp}

\title{\vspace{-0.5cm}\large{\bf{A Multi-Modal AI Copilot for Single-Cell Analysis with Instruction Following}}}

\author{
Yin Fang \textsuperscript{1,3,*},
Xinle Deng \textsuperscript{1,4,*},
Kangwei Liu \textsuperscript{1,4,*},
Ningyu Zhang \textsuperscript{1,4,$\dag$},\\
Jingyang Qian \textsuperscript{3,5},
Penghui Yang \textsuperscript{3,5},
Xiaohui Fan \textsuperscript{3,5,6,$\dag$},
Huajun Chen \textsuperscript{1,2,$\dag$},
}

\date{}

\begin{document}
\maketitle

{\renewcommand\baselinestretch{1.4}\selectfont

\noindent $^1$  College of Computer Science and Technology, Zhejiang University, Hangzhou 310027, China. \\
$^2$ ZJU-Hangzhou Global Scientific and Technological Innovation Center, Hangzhou 311200, China. \\
$^3$ College of Pharmaceutical Sciences, Zhejiang University, Hangzhou 310058, China. \\
$^4$ School of Software Technology, Zhejiang University, Ningbo 315048, China. \\
$^5$ Future Health Laboratory, Innovation Center of Yangtze River Delta, Zhejiang University, Jiaxing 314100, China. \\
$^6$ Innovation Center in Zhejiang University, State Key Laboratory of Component-Based Chinese Medicine, Hangzhou 310058, China. \\

\noindent $^*$ Equal contribution 

\noindent $\dag$ Corresponding Author: Prof. Ningyu Zhang, e-mail: \href{zhangningyu@zju.edu.cn}{\textcolor[RGB]{42, 97, 187}{\uline{zhangningyu@zju.edu.cn}}}, Prof. Xiaohui Fan, \\e-mail: \href{fanxh@zju.edu.cn}{\textcolor[RGB]{42, 97, 187}{\uline{fanxh@zju.edu.cn}}},
and Prof. Huajun Chen, e-mail: \href{huajunsir@zju.edu.cn}{\textcolor[RGB]{42, 97, 187}{\uline{huajunsir@zju.edu.cn}}} 

\par}

\renewcommand{\figurename}{Fig.}

\flushbottom

\normalsize

\section*{Abstract}

{\renewcommand\baselinestretch{1.3}\selectfont

Large language models excel at interpreting complex natural language instructions, enabling them to perform a wide range of tasks.
In the life sciences, single-cell RNA sequencing (scRNA-seq) data serves as the ``language of cellular biology'', capturing intricate gene expression patterns at the single-cell level. However, interacting with this ``language'' through conventional tools is often inefficient and unintuitive, posing challenges for researchers.
To address these limitations, we present {\ours}, a multi-modal AI copilot that leverages natural language as a medium for more direct and flexible single-cell analysis.
We construct a comprehensive multi-modal instruction dataset that pairs text-based instructions with scRNA-seq profiles from diverse tissues and species.
Building on this, we develop a multi-modal cell language architecture capable of simultaneously interpreting and processing both modalities.
{\ours} empowers researchers to accomplish critical tasks—such as cell type annotation, conditional pseudo-cell generation, and drug sensitivity prediction—using straightforward natural language commands. 
Extensive evaluations demonstrate that {\ours} consistently meets or exceeds the performance of existing single-cell foundation models, while adapting to diverse experimental conditions.  
More importantly, {\ours} provides an accessible and intuitive tool for exploring complex single-cell data, lowering technical barriers and enabling deeper biological insights.

\par}

\bigskip
\bigskip

\newpage

\section*{Main}

Advances in artificial intelligence (AI), particularly large language models (LLMs) such as GPT-4~\cite{gpt4}, PaLM~\cite{DBLP:journals/jmlr/ChowdheryNDBMRBCSGSSTMRBTSPRDHPBAI23}, LLaMA~\cite{DBLP:journals/corr/abs-2302-13971}, and Claude~\cite{claude}, have transformed natural language into a powerful medium for managing real-world tasks~\cite{DBLP:conf/nips/Ouyang0JAWMZASR22,DBLP:conf/iclr/SanhWRBSACSRDBX22}. 
Similarly, in the life sciences, single-cell RNA sequencing (scRNA-seq) data serves as a ``language of cellular biology'', encoding diverse gene expression patterns. Like grammar and vocabulary in natural language, scRNA-seq data reveals the complexity of biological systems, with each cell type and state presenting a unique expression~\cite{brazma2000gene,plass2018cell,cao2019single}.

To unlock the potential of this ``life science language'', researchers have increasingly turned to natural language processing (NLP) technologies to analyze single-cell gene expression data. These approaches typically rely on extensive gene expression datasets~\cite{barrett2012ncbi,regev2017human} for pre-training, followed by fine-tuning for specific downstream tasks~\cite{yang2022scbert,theodoris2023transfer,cui2024scgpt,hao2024large}. While effective, such methods often demand significant domain expertise and face challenges in adaptability. 
As an alternative, recent efforts have reformulated single-cell gene expression data into sequences of gene names ranked by expression levels, enabling language models to process both task-specific instructions and gene name lists directly~\cite{levine2023cell2sentence,hou2023reference}. However, focusing exclusively on top-expressed genes overlooks important information from lower-expressed ones, while incorporating all genes inflates context length and computational demands. Furthermore, converting quantitative gene expression data into text risks losing essential numerical precision.

To address these limitations, it is crucial to develop an approach that bridges the structured numerical nature of single-cell data with the expressive flexibility of natural language. 
Drawing inspiration from how humans integrate multiple sensory inputs to enhance comprehension, we propose {\ours}, a multi-modal AI copilot specifically designed for single-cell analysis. It interprets numerical single-cell data and natural language instructions, as well as generates outputs in either modality, effectively bridging the gap between these distinct data types.

First, we construct a multi-modal single-cell instruction dataset that unifies essential single-cell analysis tasks into a cohesive collection. Each cell in the dataset is represented by its gene expression profile and enriched with biological attributes specifying species, tissue, sequencing protocol, and other pertinent biological contexts, all organized into natural language instructions. To equip our AI copilot with fundamental conversational abilities, we incorporate traits reflecting diverse personalities, motivations, and proficiency levels. Using LLMs, we simulate various communication styles, enabling the copilot to adapt to diverse research contexts and expertise levels.

Second, building upon this dataset, we develop a multi-modal cell language model capable of harmonizing single-cell gene expression data with textual information. The architecture includes a Q-Former module for embedding gene expression profiles, a backbone pre-trained language model (LM) for robust textual processing, and a cell reconstruction block for generating detailed gene expression profiles. Through instruction tuning, the model gains domain-specific expertise in single-cell analysis, enabling it to adeptly handle interleaved biological and textual data. This design allows {\ours} to handle diverse input and output formats, unifying tasks of cell comprehension and generation.

Finally, we thoroughly evaluate {\ours} across a range of single-cell analysis tasks, demonstrating robustness to varied instruction styles and its ability to produce accurate, relevant outputs. Extensive experiments highlight its capacity to uncover biological insights and validate the necessity of each architectural component. {\ours} extends instruction-following capabilities to the cell-language multi-modal space, laying the groundwork for advancing single-cell research.

\section*{Results}

\subsection*{Overview of InstructCell}

\begin{figure*}[!th] 
  \vspace{-0.5cm}
  \centering
  \includegraphics[width=1\linewidth]{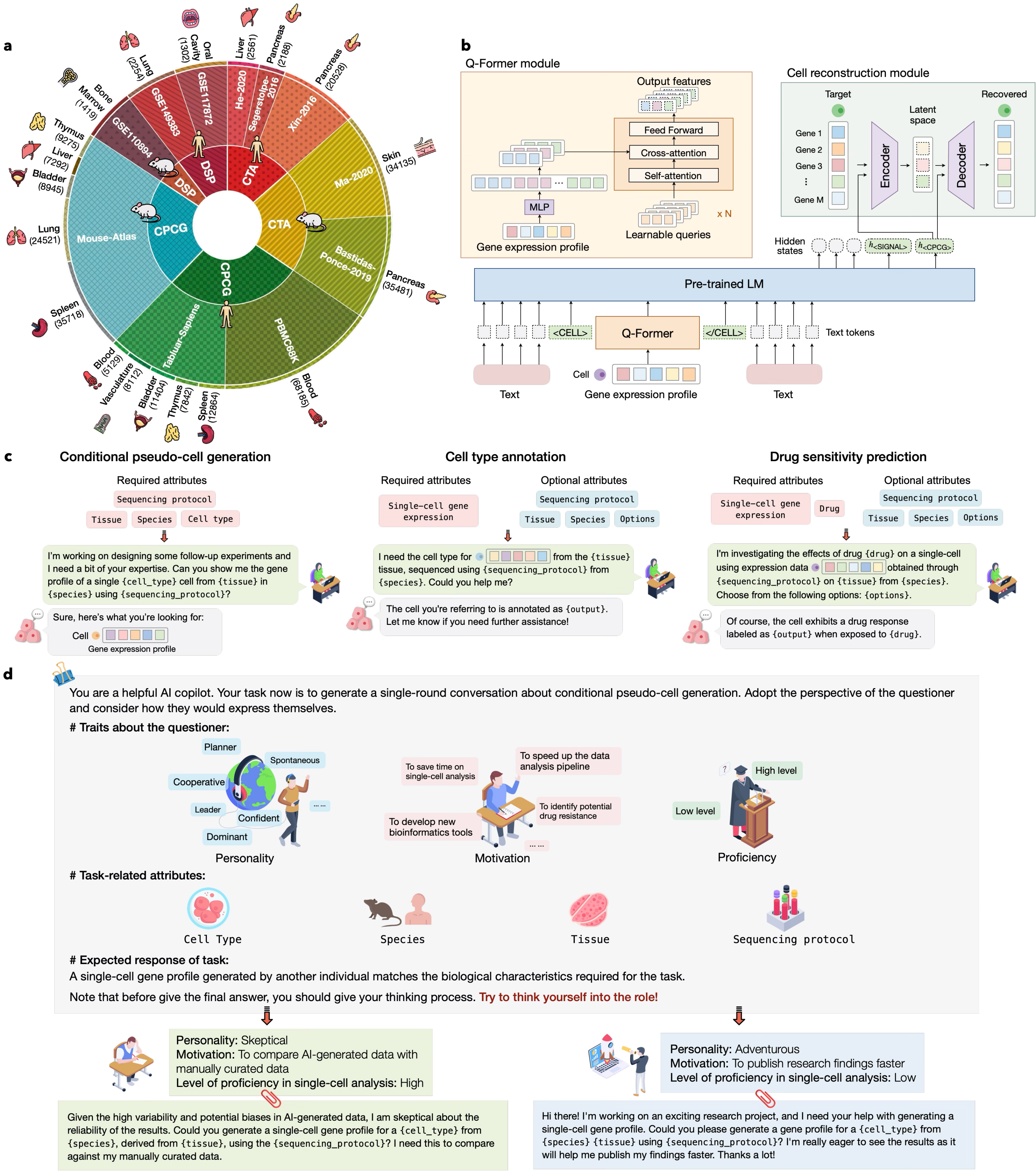}
  \caption{
  \small
  \textbf{Overview of {\ours}.}
  \textbf{a,} Summary of incorporated single-cell data. {\ours} incorporates 299,155 scRNA-seq samples from human and mouse origins, spanning multiple organs. \textit{CPCG} denotes \textit{Conditional Pseudo-cell Generation}, \textit{CTA} denotes \textit{Cell Type Annotation}, and \textit{DSP} denotes \textit{Drug Sensitivity Prediction}.
  \textbf{b,} Architecture of the multi-modal cell language model. The model processes both text and single-cell data via three primary components: a Q-Former to capture single-cell gene expression knowledge, a pre-trained LM as the backbone, and a cell reconstruction module for generating single-cell gene expression profiles.
  \textbf{c,} Construction of multi-modal single-cell instruction data. Complete instruction-response pairs are formed by combining required and optional attributes from text and single-cell modalities.
  \textbf{d,} Simulation of diverse communication styles. LLMs generate chat templates with varying traits (personality, motivation, and proficiency) to produce instructions that convey task-related information in different communication styles.
  }
  \vspace{-1cm}
  \label{fig:overview}
\end{figure*}

In this paper, we propose {\ours}, a multi-modal AI copilot specifically designed for single-cell analysis. The development of {\ours} involves two critical components: (1) constructing multi-modal single-cell instruction data and (2) training a multi-modal cell LM. An overview of {\ours} is illustrated in Fig.~\ref{fig:overview}.

\paragraph{Multi-modal single-cell instruction data construction}

{\ours} aims to advance single-cell analysis by leveraging natural language as an interactive medium. Current models typically rely on a single modality, focusing either on single-cell data alone~\cite{cui2024scgpt,yang2022scbert,theodoris2023transfer}, text data alone~\cite{levine2023cell2sentence,hou2023reference}, or deriving cell embeddings directly from text~\cite{chen2023genept,liu2023scelmo}. 
Consequently, available datasets are predominantly rooted in scRNA-seq data~\cite{barrett2012ncbi,regev2017human} or textual representations derived from it~\cite{levine2023cell2sentence,chen2023genept}. 
Thus, our primary goal is to construct a multi-modal single-cell instruction dataset that enables the model to comprehend and process both the ``language'' of life sciences and human language effectively.

As shown in Fig.~\ref{fig:overview}(a), we focus on two species (human and mouse) and collect scRNA-seq datasets from multiple tissues. These datasets are organized into gene expression count matrices, where rows represent individual cells, columns represent genes, and entries indicate gene expression levels. To provide essential biological context, we also record attributes such as tissue, species, cell type, and sequencing protocol for each dataset.

Using these biological datasets, we create complete instruction-response pairs in natural language. Since different tasks emphasize distinct aspects of the data, the corresponding instructions and responses draw on varying subsets of attributes. As illustrated in Fig.~\ref{fig:overview}(c), our multi-modal single-cell instruction dataset covers three key tasks: \textit{Conditional Pseudo-cell Generation} (\textit{CPCG}), which focuses on generating gene expression profiles tailored to specific cell types and biological conditions; \textit{Cell Type Annotation} (\textit{CTA}), which seeks to accurately classify cells based on their gene expression profiles; and \textit{Drug Sensitivity Prediction} (\textit{DSP}), which aims to predict how cells respond to various drugs.

Each task requires specific biological attributes (highlighted in pink in Fig.~\ref{fig:overview}(c)), which are transformed into natural language instructions using GPT-4o~\cite{gpt4o}. To enrich biological context, optional attributes (blue tags) provide additional background without affecting the instruction’s core completeness. 
While LLMs can jointly generate instructions and responses, this often results in brief, less informative outputs~\cite{DBLP:conf/emnlp/DingCXQHL0Z23}. To address this, we employ two separate LLMs: the first generates instruction templates, and the second produces response templates based on the output of the first. We produce two types of response templates: interleaved formats that combine dialogue text with answer labels for interactive applications, and simplified formats that provide only answer labels for concise outputs. These two formats support the training of \textit{chat} and \textit{instruct} versions of {\ours}, accommodating both interactive dialogues and task-specific outputs.

To enable {\ours} to comprehend instructions from users with diverse cultural and linguistic backgrounds, we use LLMs to simulate different language styles, specialized terminology usage, and question structures. Specifically, we define three main traits (Fig.~\ref{fig:overview}(d)): personality, which influences speaking styles; motivation, which adds richness to the dialogue context; and proficiency, which determines the level of technical jargon used. By combining these traits with task-related attributes, we generate a wide range of instruction templates that enhance the model’s ability to handle complex queries, adapt to real-world scenarios, and personalize user interactions.

\paragraph{Multi-modal cell language model}

To enable the model to handle both text and single-cell data modalities concurrently, {\ours} is built upon a multi-modal LM architecture designed to facilitate cross-modal knowledge sharing and enhance its ability to process diverse input types. 
As illustrated in Fig.~\ref{fig:overview}(d), the architecture comprises three main components: 
(1) A Q-Former~\cite{DBLP:conf/icml/0008LSH23} that extracts and encodes features from single-cell gene expression data;
(2) A pre-trained LM~\cite{radford2019language,DBLP:conf/acl/LewisLGGMLSZ20,DBLP:journals/jmlr/RaffelSRLNMZLL20} serving as the backbone for text processing;
(3) A cell reconstruction module that enables the generation of single-cell gene expression profiles.

To differentiate single-cell data from natural language, we introduce special tokens \texttt{<CELL>} and \texttt{</CELL>} into the tokenization process.  These tokens delineate single-cell data segments, ensuring the model correctly identifies and processes this modality~\cite{wu2024towards,jiang2024mantis}.

We employ the Q-Former to extract cell features from the gene expression data. The Q-Former includes a transformer submodule with a set of learnable query embeddings that capture information through self-attention and cross-attention layers. The output is a collection of encoded cell features. These features, along with the processed textual data, are fed into the pre-trained LM’s transformer blocks. At this stage, both the cell features and text have been transformed into vector representations, eliminating the need for further embeddings.

Once the model encodes both modalities, it produces a sequence of hidden states that represent their contextual relationships. Attention mechanisms guide the model to relevant segments of the input and previously generated outputs. Generation continues until reaching an end-of-text token or a special \texttt{<SIGNAL>} token~\cite{wu2023next,DBLP:conf/nips/KohFS23} (decoded from the hidden state vector $\boldsymbol{h}_{\texttt{<SIGNAL>}}$), which indicates the need to produce single-cell data next. If no single-cell generation is required, the \texttt{<SIGNAL>} token does not appear. For single-cell generation tasks, the model learns to output \texttt{<SIGNAL>} at the appropriate time, signaling the transition to cell data generation.

After producing the \texttt{<SIGNAL>} token, it and the previously generated text are fed back into the LM. Through attention, the model obtains a final hidden state vector $\boldsymbol{h}_{\texttt{<CPCG>}}$ that integrates both the instruction information and the preceding textual context. To generate the single-cell gene expression profile, we employ a cell reconstruction module based on a conditional variational autoencoder (CVAE)~\cite{DBLP:conf/nips/SohnLY15}. Given the target gene expression profile and $\boldsymbol{h}_{\texttt{<CPCG>}}$, the CVAE encoder and decoder guide the model to reconstruct the original profile. This approach allows the model to accurately reproduce observed patterns in the data while navigating the latent space smoothly, producing new, biologically coherent instances.
During inference, we retain only the cell decoder block. A single-cell gene expression profile is generated by sampling a latent vector from an isotropic Gaussian distribution conditioned on $\boldsymbol{h}_{\texttt{<CPCG>}}$.

In summary, this multi-modal cell language model flexibly processes and generates essential biological information for single-cell analysis tasks. By bridging the language of life sciences and human communication, it lowers knowledge barriers, accelerating the acquisition of accurate and robust biological insights.

\subsection*{{\ours} enables conditional pseudo-cell generation}

\begin{figure*}[!th] 
  \centering
  \includegraphics[width=1\linewidth]{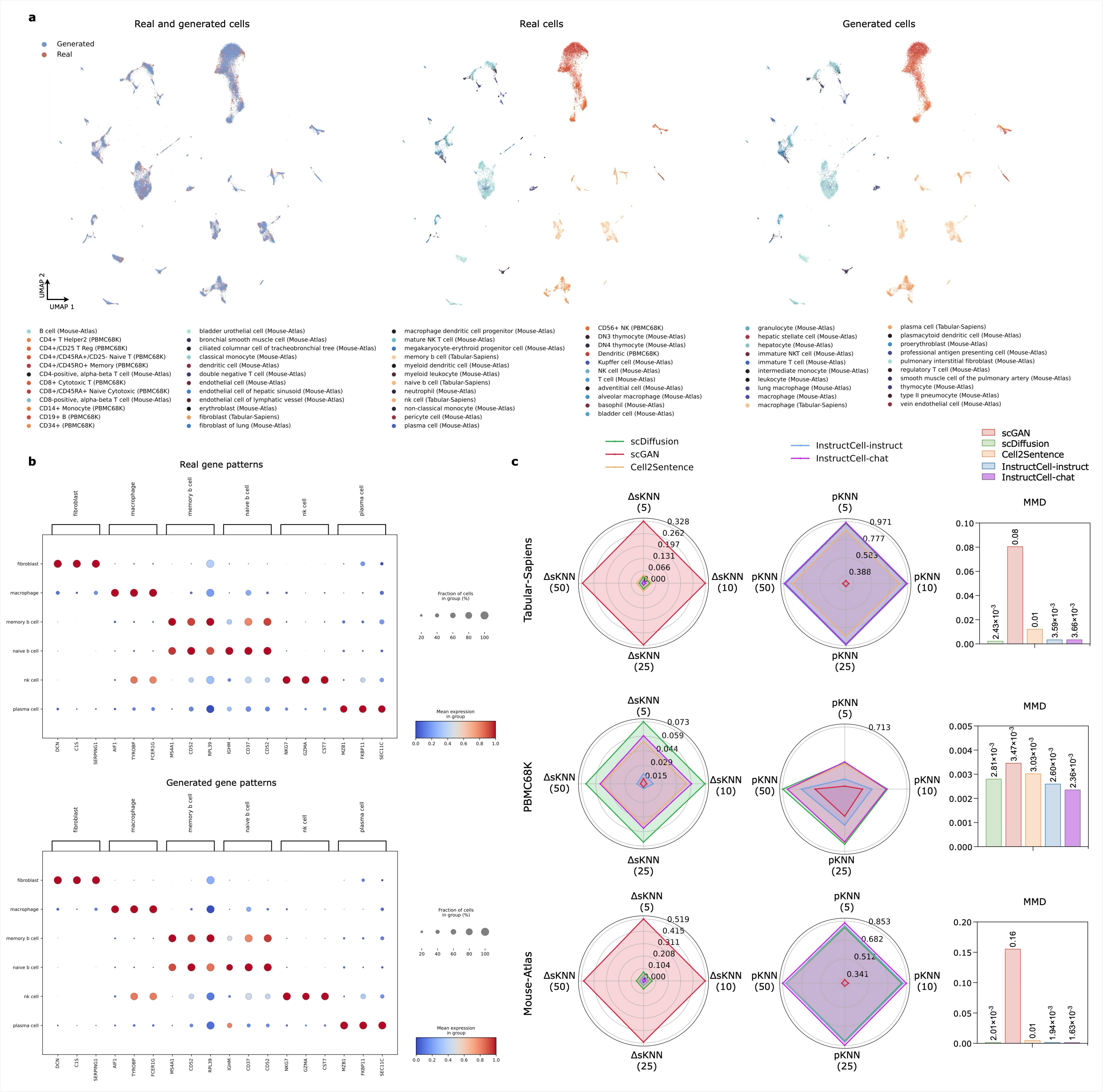}
  \caption{
  \small
  \textbf{Conditional pseudo-cell generation results by {\ours}.}
  \textbf{a,} UMAP visualizations of real and generated cells. The left plot shows the overlap between real and generated cells. The middle and right plots display real and generated cells, respectively, with distinct colors indicating different cell types.
\textbf{b,} Dot plots of gene expression patterns derived from real (top) and generated (bottom) cells. Based on the test set from Tabular-Sapiens, we use Welch's $t$-test to identify top three significant genes for each cell type and display them along x-axis. Cell types are arranged along y-axis. The size of each dot indicates the proportion of single cells within the corresponding cell type that express the gene, while the color of the dot represents the mean expression level of the gene within that cell type. The results of the remaining two datasets are available in Fig.\ref{fig:bubble_plot}.
  \textbf{c,} Quantitative evaluation of cell generation performance across four datasets. A lower $\triangle$sKNN value indicates better structural alignment, a higher pKNN value reflects improved positional correspondence, and a lower MMD value denotes a more accurate approximation of the global data distribution.
  }
  \label{fig:cell_generation}
  \vspace{-0.5cm}
\end{figure*}

\textit{CPCG} is a task that requires the model to generate a single-cell gene expression profile matching specified conditions, such as cell type, tissue, and species, as illustrated in Fig.~\ref{fig:overview}(c). 
To explore the generative modeling capability of {\ours}, experiments are conducted using scRNA-seq data from 9 tissues —bladder, blood, liver, lung, spleen, thymus, and vasculature—collected from humans and mice. The datasets utilized in these experiments include Tabular-Sapiens, Mouse-Atlas, and PBMC68K.
First, we directly provide the generation conditions to {\ours} in textual form, and it generates cells based on these conditions.
Fig.~\ref{fig:cell_generation}(a) visualizes the comparison between the real cells' gene expression profiles and those generated by {\ours}. 
The generated cell distributions closely resemble those of real cells, even accurately replicating some outliers.
Notably, the instruction-response templates used for generation are not included in the training phase, demonstrating that the model is robust to unseen input templates.
This suggests that {\ours} effectively generalizes to new instructions by capturing the underlying biological patterns rather than overfitting to specific training templates.
Additionally, we observe that the UMAP visualizations of generated cells exhibit slightly higher clustering compared to real cells. While this clustering might suggest a slight oversimplification of the inherent biological variability, it reflects the model’s ability to faithfully capture the overall structure of the data and preserve key features in the reduced-dimensional space.

Furthermore, we investigate the model’s ability to learn single-cell expression patterns specific to each cell type. As depicted in Fig.~\ref{fig:cell_generation}(b), each dot plot organizes genes on the x-axis and different cell types on the y-axis, with each cell type being grouped by its top three most significant genes.  
To identify the key expressive genes for each cell type, we compare individual cells to the average expression levels of other cells within the same cell type. Specifically, for each cell, we consider its gene expression levels and compare them to the average gene expression levels across all other cells of the same cell type.
We conduct Welch's $t$-test~\cite{welch1947generalization} for each gene, comparing the expression level in the individual cell to the average expression in the other cells. 
A significant result indicates that the gene is particularly expressive or distinctive in that cell compared to others of the same type. We systematically identify the top three most significant genes for each cell type, recognizing these as the key expressive genes.
For the cells generated by the model, we directly display the expression proportions and average expression levels of these significant genes.
By closely examining these patterns, we find that the generated cells faithfully replicate the gene expression profiles of real cells, particularly in terms of gene-specific expression levels and cell-type-specific patterns. These results demonstrate the model’s capacity to generate biologically plausible gene expression profiles that preserve intricate gene-level and cell‐type‐specific patterns critical for realistic simulations.

We further compare {\ours} against existing cell generation methods across three dimensions, as illustrated in Fig.~\ref{fig:cell_generation}(c). 
As detailed in Methods, $\triangle$sKNN quantifies the structural divergence between generated and real data by examining label consistency within cell clusters in the reduced-dimensional space, while pKNN evaluates positional alignment by assessing how accurately generated cells match specific cell types.
We report the average $\triangle$sKNN and pKNN values for K of 5, 10, 25, and 50 across all datasets in the radar charts. The radial axis range is set according to the maximum value observed in the data, providing a clear visual reference. All radar charts in this paper follow this convention, and no further explanation will be provided for subsequent figures.
Both the \textit{instruct} and \textit{chat} versions of {\ours} closely align with real cells, effectively replicating biological structures and maintaining accurate cell positioning. The MMD metric, which measures the overall similarity between generated and real cells, shows that both versions of {\ours} achieve significantly lower MMD values compared to baseline models.  

Among the baseline methods, scDiffusion~\cite{luo2024scdiffusion} and scGAN~\cite{marouf2020realistic} focus solely on cell generation. Unlike the end-to-end training strategy of {\ours}, scDiffusion requires separately trained modules, which introduces additional complexity. scGAN, on the other hand, is known for training difficulties, mode collapse, and instability. Moreover, scDiffusion and scGAN produce continuous gene expression values, whereas scRNA-seq data are inherently discrete, and some downstream tasks require discrete inputs. Similarly, the gene expression profiles generated by Cell2Sentence~\cite{levine2023cell2sentence} are reconstructed in expression space through normalization and log1p-transformation, losing their original discrete characteristics.
In contrast, {\ours} preserves the discrete nature of gene expression profiles by treating them as a distinct modality, maintaining precise numerical information and capturing low-expression genes. This approach makes {\ours} more scalable and flexible, offering distinct advantages in handling complex datasets and adapting to diverse experimental scenarios.

Overall, these findings suggest that {\ours} effectively captures the essential characteristics of different cell types, ensuring robust performance in cell simulation tasks. Its intuitive design allows the direct generation of single-cell gene profiles from verbal descriptions of conditions, providing a practical and adaptable tool for producing specific cellular data in controlled experimental contexts.

\subsection*{{\ours} boosts the performance of cell type annotation}

\begin{figure*}[!t] 
  \centering
  \includegraphics[width=0.9\linewidth]{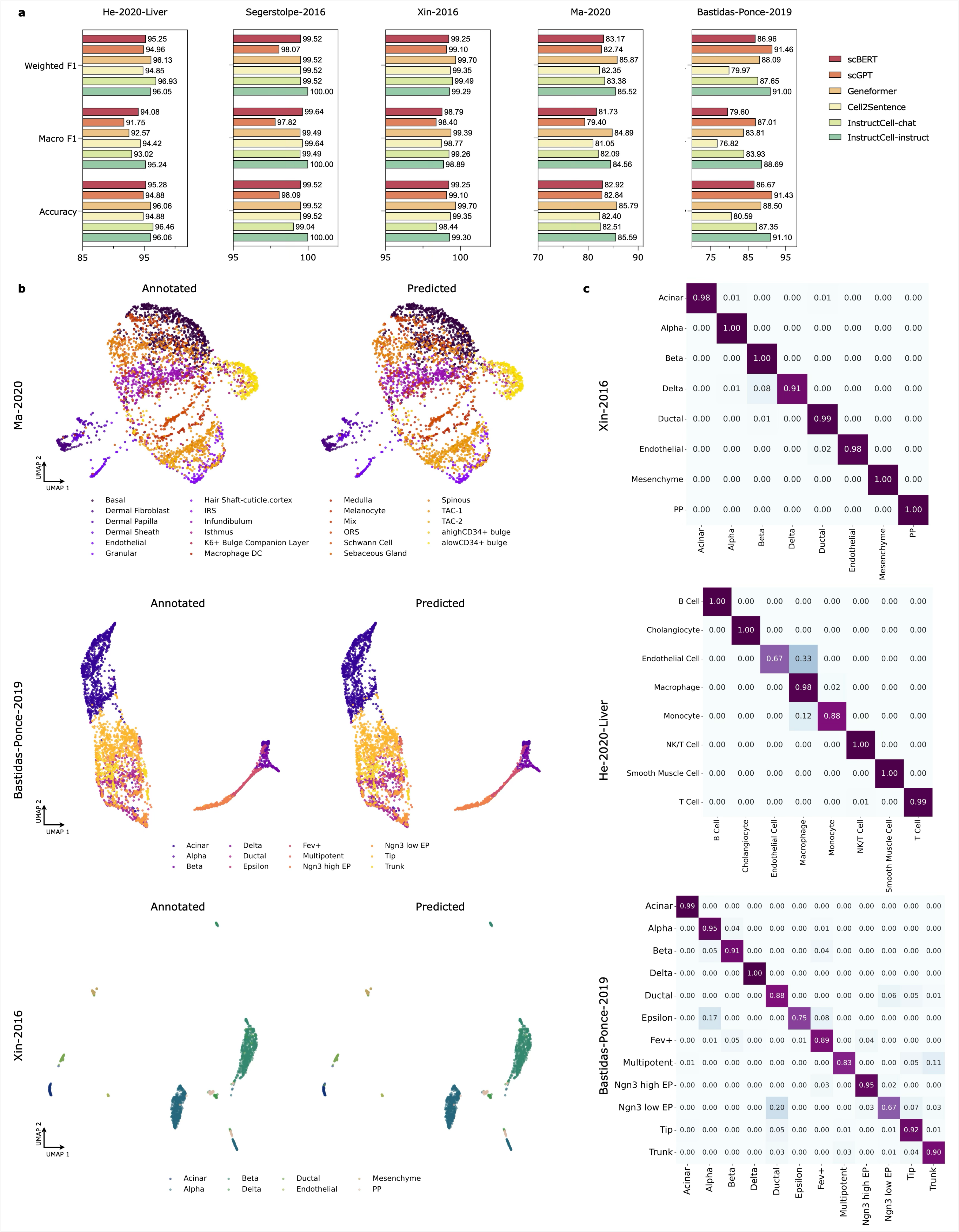}
  \caption{
  \small
  \textbf{Cell type annotation results by {\ours}.}
  \textbf{a,} Evaluation of {\ours}'s \textit{CTA} performance across human heart, liver, pancreas, and mouse skin and pancreas datasets. Performance is quantified using weighted F1, macro F1, and accuracy metrics, with different colors representing different models.
  \textbf{b,} UMAP visualization of three different datasets. The left panel is colored by expert-annotated cell types from the original research, and the right panel is colored by {\ours} prediction results.
  \textbf{c,} Confusion matrices between predicted cell types and actual annotations for the three datasets. Darker shades denote a higher frequency of agreement between the model's predictions and the actual cell type annotations.
  }
  \vspace{-0.5cm}
  \label{fig:cell_type}
\end{figure*}

Annotating cell types based on scRNA-seq data is a fundamental task in single-cell analysis. {\ours} innovates in \textit{CTA} by reframing it as a single-cell question-answering task. Given an interleaved sequence of texts and gene expression profile, {\ours} generates the textual response that conveys its prediction regarding the cell type (Fig.~\ref{fig:overview}(c)). 
To assess whether the model can accurately differentiate between various cell types, we first benchmark it across 5 scRNA-seq datasets—Xin-2016, Segerstolpe-2016, He-2020-Liver, Ma-2020, and Bastidas-Ponce-2019. These datasets span 3 major organs/tissues (liver, pancreas, and skin) in humans and mice, covering more than 40 cell types and over 94,000 cells, utilizing mainstream scRNA-seq protocols such as Smart-seq2 and HiSeq X Ten sequencing. The datasets chosen showcase a range in both data size and complexity, as depicted in Fig.~\ref{fig:overview}(a) and Fig.~\ref{fig:cell_type}(a). 

From Fig.~\ref{fig:cell_type}(a), we observe that {\ours} performs on par with, or even surpasses, foundation models (scBERT, scGPT, and Geneformer) across nearly all datasets, despite not relying on large-scale unlabeled pre-training. This indicates that {\ours} remains robust and reliable when handling a diverse range of single-cell data. In contrast, foundation models often require extensive fine-tuning on each dataset for multiple tasks, resulting in greater training complexity, more cumbersome deployment, and a substantially larger number of parameters.
While Cell2Sentence simplifies representation, it may omit valuable quantitative details and exclude potentially informative, lower-expressed genes. By contrast, {\ours} preserves a broader spectrum of gene expression information.
This design enables the model to better capture the biological complexity of single-cell data, enhancing its ability to align with natural language instructions and deliver more accurate outputs.

Fig.~\ref{fig:cell_type}(b) shows that the cell types predicted by {\ours} are highly consistent with the original cell types. This is particularly evident for similar cell types with overlapping characteristics, where the model demonstrates an ability to distinguish subtle differences with minimal errors.
Fig.~\ref{fig:cell_type}(c) indicates that {\ours} achieves high accuracy (>0.9) for most cell types displayed in the confusion matrix.   
This consistent performance across diverse cell types demonstrates the model’s capacity to accurately interpret single-cell data and predict cell types, even in the presence of interleaved text, without requiring extensive pre-training or dataset-specific fine-tuning.
The confusion matrices for the remaining datasets are presented in Fig.~\ref{fig:confusion_matrix}.
Beyond its consistent performance across diverse cell types, {\ours} demonstrates adaptability in handling data from different species, ensuring accurate predictions across varied biological contexts. 
This cross-species compatibility enhances its potential as a practical tool for single-cell data analysis, supporting tasks such as comparative studies and integrated research pipelines.

\subsection*{{\ours} enhances precision of drug sensitivity prediction}

\begin{figure*}[!ht] 
  \centering
  \includegraphics[width=0.82\linewidth]{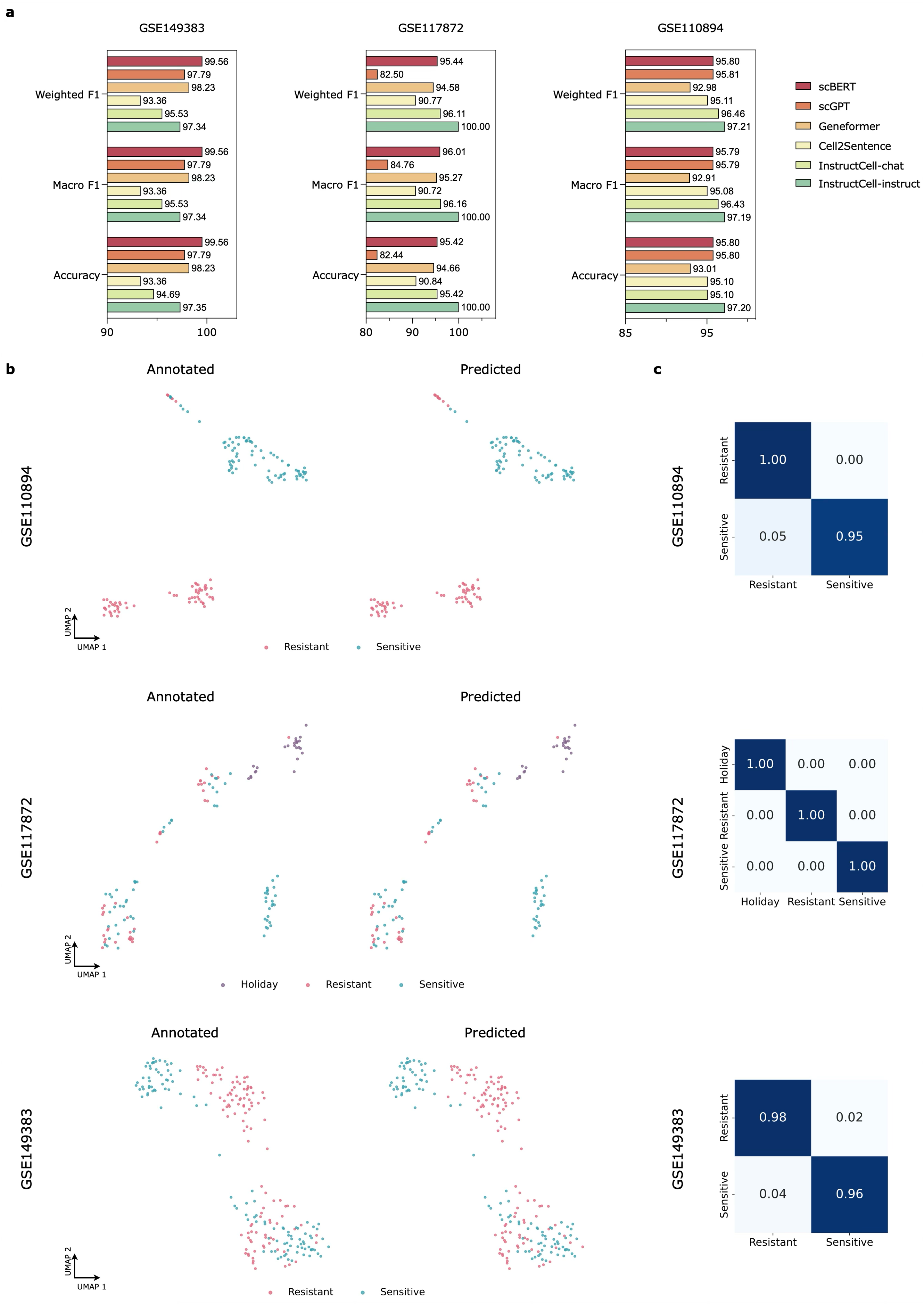}
  \caption{
  \small
  \textbf{Drug sensitivity prediction results by {\ours}.}
  \textbf{a,} Evaluation of {\ours}'s \textit{CTA} performance across human oral, lung, and mouse bone datasets. Performance is quantified using weighted F1, macro F1, and accuracy metrics, with different colors representing different models.
  \textbf{b,} UMAP visualization of the three datasets, with cells colored by drug sensitivity labels (sensitive, resistant, and holiday) for both expert-annotated results and {\ours} predictions. 
  \textbf{c,} Confusion matrices between predicted cell types and actual annotations for the three datasets. Darker shades denote a higher frequency of agreement between the model's predictions and the actual drug sensitivity annotations.
  }
  \label{fig:drug_sensitivity}
  \vspace{-0.6cm}
\end{figure*}

As illustrated in Fig.~\ref{fig:overview}(c), \textit{DSP} involves providing the model with drug information and single-cell gene expression data, allowing it to predict whether a cell is resistant or sensitive to a given drug. For our experiments, we gathered scRNA-seq data from three organs, including datasets from humans (GSE149383 and GSE117872) and mice (GSE110894), paired with drug sensitivity information. Notably, the GSE117872 dataset includes an additional category, labeled `holiday', which refers to observations made during off-treatment periods.

Fig.~\ref{fig:drug_sensitivity}(a) shows that {\ours} achieves superior performance across all three evaluation metrics on the GSE117872 and GSE110894 datasets, while on the GSE149383 dataset, it delivers results comparable to single-cell foundation models. These results demonstrate the model’s ability to predict single-cell responses to different drugs based on textual descriptions of conditions and gene expression data. 

Fig.~\ref{fig:drug_sensitivity}(b) presents a UMAP visualization comparing annotated and predicted single cells with different labels. The visualization demonstrates that {\ours} effectively distinguishes these categories, accurately capturing the separation between drug-sensitive and drug-resistant cells while appropriately positioning holiday cells. This suggests that the model successfully reflects biological differences in drug responses within the reduced-dimensional space. 

Fig.~\ref{fig:drug_sensitivity}(c) presents a detailed confusion matrix for each label, showing that {\ours} achieves high accuracy levels (>0.95) across different species and tissues. This consistent performance highlights the model’s robustness in integrating single-cell gene expression data with drug information, making it a reliable tool for drug response prediction across diverse biological contexts.

\subsection*{{\ours} demonstrates robustness to varied instruction styles}

\begin{figure*}[!t] 
  \centering
  \includegraphics[width=1\linewidth]{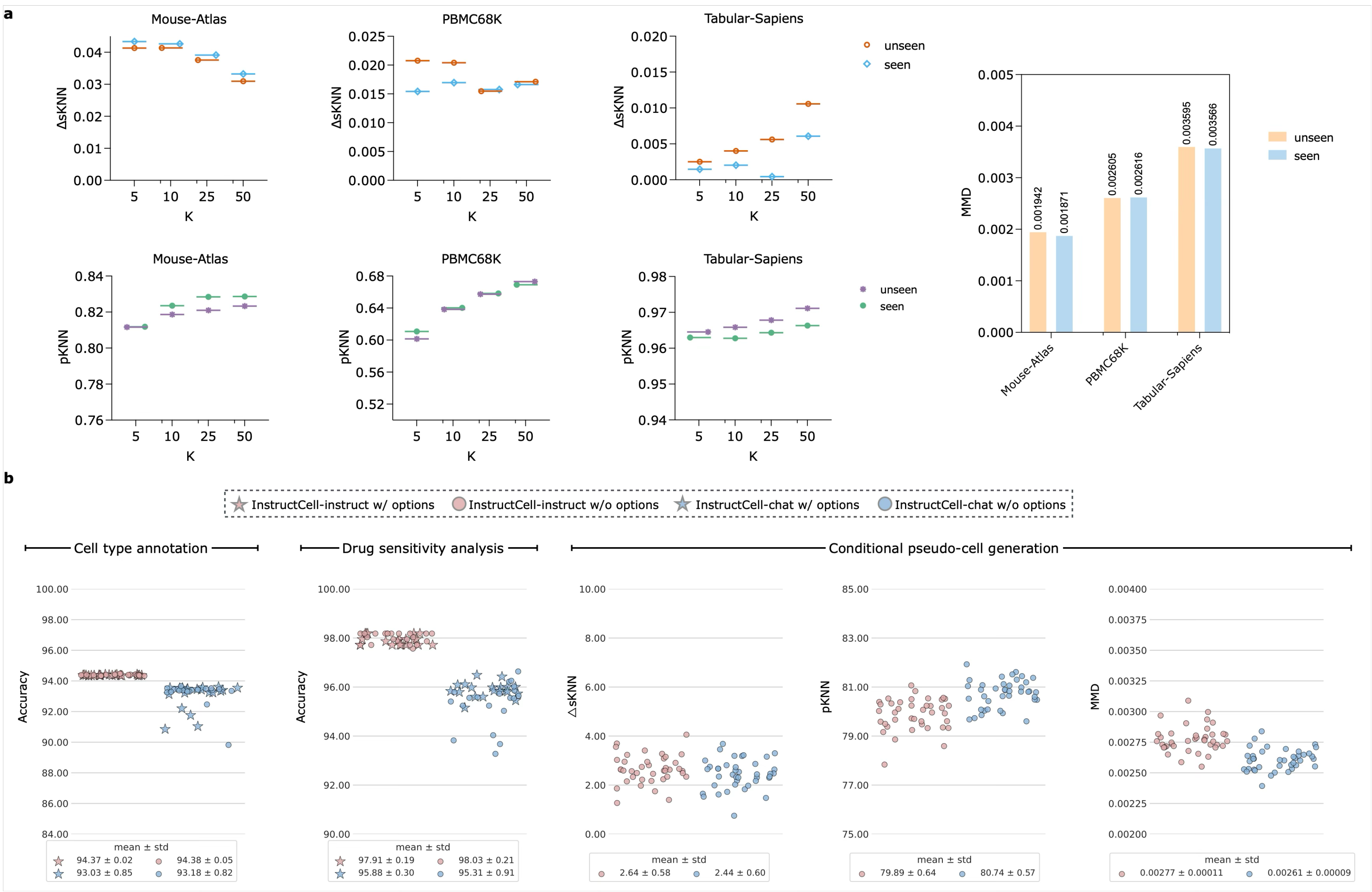}
  \caption{
  \small
  \textbf{Robustness of {\ours}.}
  \textbf{a,} Quantitative comparison of the \textit{CPCG} task under seen and unseen instruction templates.  
  Results are shown for $\triangle$sKNN and pKNN metrics at varying numbers of neighbors $K$, as well as for MMD. Different colors denote whether the instruction templates are seen or unseen.
  \textbf{b,} Average performance of {\ours} under instruct and chat modes across each task. On the left side (classification tasks), the shape of each scatter point indicates whether options are provided or not, while the color distinguishes model versions. Each configuration includes 40 scatter points (20 with options and 20 without). On the right side (generative task), different colors represent different model versions.
  }
  \label{fig:robustness}
  \vspace{-0.3cm}
\end{figure*}

As a tool designed to assist researchers, {\ours} must effectively understand and address questions phrased differently. Fig.~\ref{fig:robustness} provides quantitative evidence for this adaptability.
To assess how the model reacts to unfamiliar instruction templates, we first test it on the \textit{CPCG} task, which is more demanding than the two classification tasks. We measure several metrics on cells generated using both unseen and seen templates across three datasets. As shown in Fig.~\ref{fig:robustness}(a), the {$\triangle$}sKNN, pKNN, and MMD values remain similar in both conditions, indicating that {\ours} maintains consistent performance regardless of prior exposure to the instruction templates. This implies that {\ours} preserves both detailed structural features and overall distributional patterns, allowing it to adapt effectively to different researchers’ phrasing styles.

To further investigate this adaptability, we evaluate both the \textit{instruct} and \textit{chat} versions of {\ours} on all tasks, as shown in Fig.~\ref{fig:robustness}(b). The first two panels focus on classification scenarios. In some templates, we provide multiple-choice options; however, whether these options are present or not has almost no impact on accuracy. This flexibility is beneficial, as it means the model does not need pre-defined options to achieve high performance, helping keep inputs short and avoiding constraints that could slow inference due to quadratic complexity. That said, the \textit{instruct} version proves slightly more robust than \textit{chat}, likely because it is optimized for direct, goal-oriented instructions, thus maintaining stable performance across varied input formats. In contrast, the \textit{chat} variant, designed for open-ended exchanges, may be more sensitive to subtle stylistic shifts.

Finally, the last panel in Fig.~\ref{fig:robustness}(b) presents average results for the \textit{CPCG} task over three datasets. Both \textit{instruct} and \textit{chat} perform comparably, with consistently low standard deviations, indicating that either approach can reliably produce realistic pseudo-cell data. This outcome suggests that the model’s internal representation of biological patterns is sufficiently robust that differences in instruction style exert minimal influence on generation quality.

\subsection*{{\ours} identifies biologically significant marker genes}

\begin{figure*}[!ht] 
  \centering
  \includegraphics[width=1\linewidth]{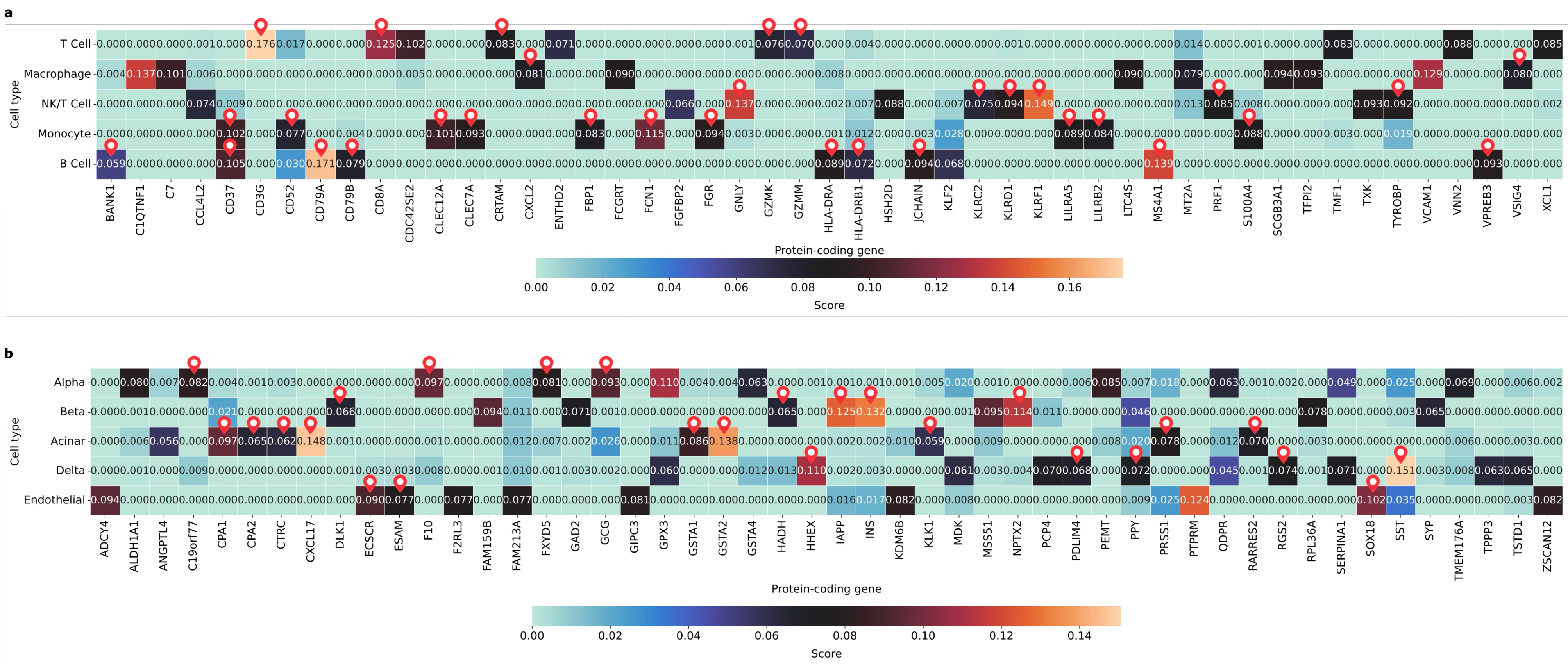}
  \caption{
  \small
  \textbf{Top 10 significant genes identified by {\ours} for each cell type in two datasets.}
  \textbf{a, b} Heatmaps of the significant genes extracted from {\ours} by using gradient saliency-based method for (a) He-2020-Liver and (b) Xin-2016 datasets. 
  The color gradient from red to blue represents gene importance, with red indicating higher importance scores and blue indicating lower scores. Red markers in each row indicate that genes among the top 10 key genes identified by the model, are either reported as marker genes for the corresponding cell type in the CellMarker2.0 database or in recent literature.
  }
  \label{fig:markergene}
\end{figure*}

To evaluate the model’s ability to uncover biologically meaningful insights, we explore whether it can identify marker genes without prior knowledge of known cell-type markers injected into the model.
Since scRNA-seq datasets provide gene expression profiles, we employ gradient-based saliency methods to determine which genes most strongly influence the model’s predictions.
Specifically, we employ the Vanilla Gradient method~\cite{DBLP:journals/corr/SimonyanVZ13} to extract the top 10 most significant genes for each cell type, as detailed in the Methods section. We then evaluate the biological relevance of these genes by comparing them with documented marker genes in the CellMarker2.0 database and recent literature~\cite{DBLP:journals/nar/HuLXZLBCJYOLWZ23}.

Fig.\ref{fig:markergene} presents results for two datasets: (a) He-2020-Liver and (b) Xin-2016. Red markers in each row indicate that genes among the top 10 key genes identified by the model, are either reported as marker genes for the corresponding cell type in the CellMarker2.0 database or in recent literature.
From two heatmaps, we observe that the genes assigned higher scores by the model are highly likely to be documented as marker genes for their respective cell types. For most cell types, more than half of the top 10 genes identified by the model are found in the CellMarker2.0 database. 
In cases where a gene from the top 10 is not explicitly reported as a marker gene in the database, it may still be highlighted in the literature as a relevant marker. For example, HHEX is identified as a highly expressed gene for human Delta cells~\cite{muraro2016single,van2022generation}, although it is not categorized as a marker gene for human Delta cells in the database.
These findings suggest that {\ours} not only effectively identifies established marker genes but also has the potential to uncover novel marker genes.

\subsection*{{\ours} exhibits high-quality expressive capabilities}

\begin{figure*}[!t] 
  \centering
  \vspace{-0.5cm}  
  \includegraphics[width=0.9\linewidth]{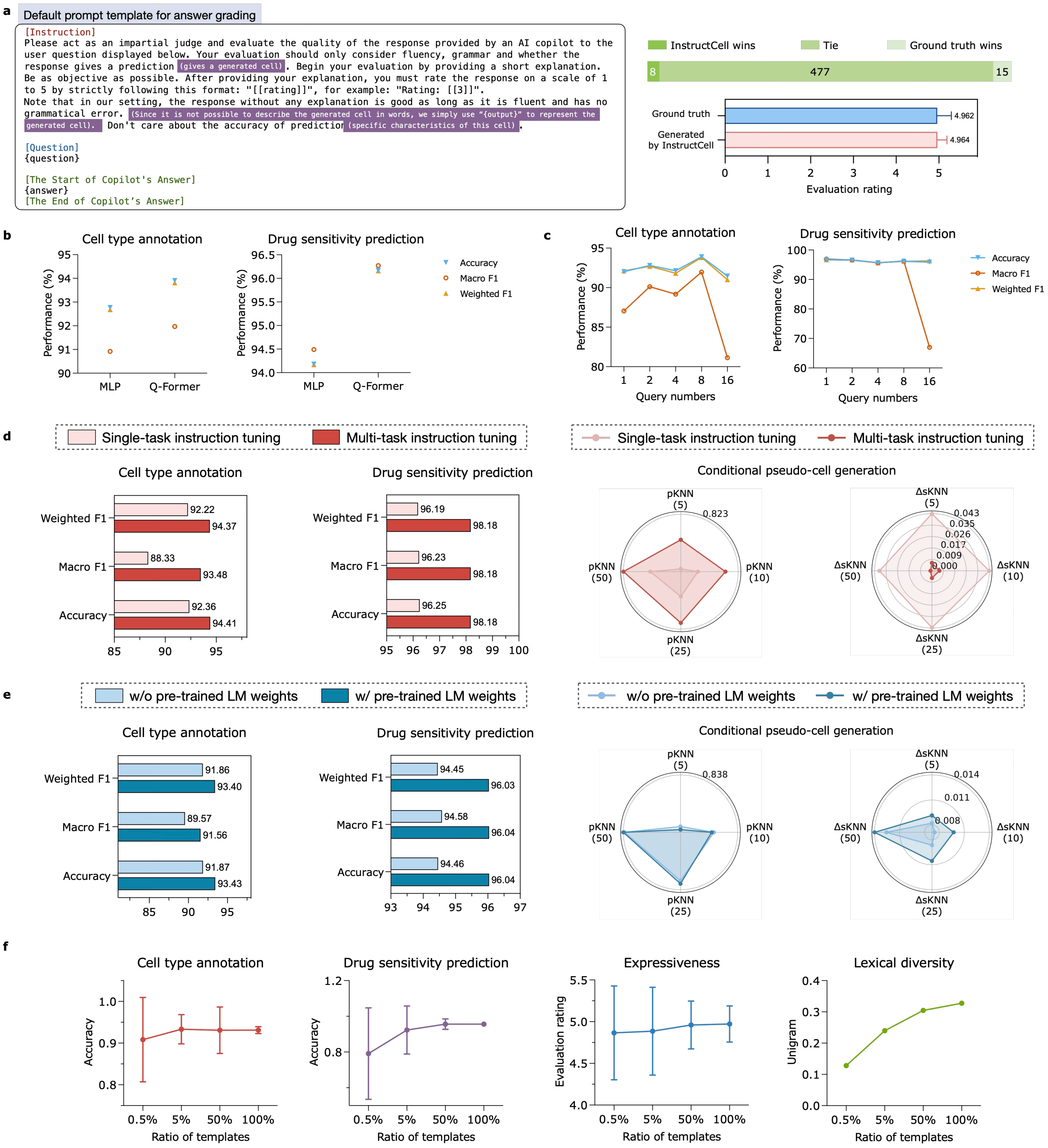}
  \caption{
  \small
  \textbf{A closer look of {\ours}.}
  \textbf{a,} Evaluation of response quality in {\ours} using the LLM-as-a-judge approach. Response quality is assessed based on fluency, grammar, and inclusion of predictive results, with Claude 3.5 Sonnet~\cite{claude} serving as an unbiased evaluator. Text highlighted in purple indicates additional content for the \textit{CPCG} task compared to the two classification tasks.
  \textbf{b,} Impact of Q-Former on model performance.
  Performance comparison between the Q-Former and a standard MLP for encoding single-cell data.
  \textbf{c,} Impact of query embedding quantity on model performance.
  Performance comparison across different numbers of query embeddings.
  \textbf{d,} Comparative performance of multi-task vs. single-task instruction tuning. For single-task instruction tuning, we divide our multi-modal instruction dataset by task type and train separate models for each specific task. We report the average metrics across all datasets for each task using the {\ours}-\textit{instruct} version.
  \textbf{e,} Comparative performance of without vs. with pre-trained LM weights. We conduct experiments using the {\ours}-\textit{chat} version to explore the impact of employing or not employing pre-trained weights on model performance. 
  \textbf{f,} Impact of varying template ratios on model outputs.
  Four \textit{chat} version models are trained on all classification datasets using multi-task instruction tuning with varying ratios of templates (0.5\%, 5\%, 50\%, and 100\% of the total templates). For the \textit{CTA} and \textit{DSP} tasks, 40 unseen instruction templates are selected for evaluation: 20 with multiple-choice options and 20 without. The mean performance and standard deviation for each template are calculated across all datasets for these two tasks. Additionally, we sample 500 unseen instruction templates and use Claude 3.5 Sonnet to score the model’s outputs for expressiveness, while unigram analysis is conducted to assess lexical diversity.
  }
  \label{fig:closer}
  \vspace{-1cm}
\end{figure*}

To maximize the scientific utility of {\ours}, it is crucial that its interactions with researchers remain both accurate and efficient. Accordingly, we assess the quality of its responses by evaluating fluency, grammatical integrity, and the inclusion of predictive results.  We randomly sample 500 entries from the multi-modal single-cell instruction templates for quality assessment.

Although manual evaluation is the gold standard for capturing human preferences, it is slow and resource-intensive. As an alternative, we adopt an `LLM-as-a-judge’ approach, using Claude 3.5 Sonnet~\cite{claude} as a surrogate evaluator. This strategy mitigates evaluation bias~\cite{DBLP:conf/nips/ZhengC00WZL0LXZ23,DBLP:journals/corr/abs-2311-09766} because Claude’s underlying architecture differs substantially from GPT-4o, which generated the original data.

Fig.~\ref{fig:closer}(a) shows the default prompt template used for these LLM-based assessments. According to the evaluation ratings, 97\% of {\ours}-generated responses meet or exceed the quality of the provided ground truth. Furthermore, the close alignment in scores between {\ours}-generated responses and the ground truth underscores the model’s ability to produce contextually accurate outputs. This competence in effective, high-quality communication suggests that {\ours} is well-positioned to serve as a tool in scientific collaborations.

\subsection*{Q-Former facilitates smooth integration of single-cell data}

To assess the effect of using the Q-Former as a feature extractor for single-cell data, we focus on classification tasks in single-cell analysis. We choose these tasks because the modifications primarily affect the input module rather than the model’s generative components for the single-cell modality. We employ the {\ours}-\textit{instruct} model for two classification tasks—\textit{CTA} and \textit{DSP}—and report the average performance across all datasets.

To determine the necessity of the Q-Former, we replace it with a standard Multi-Layer Perceptron (MLP) and compare the results. As shown in Fig.~\ref{fig:closer}(b), using a conventional MLP to encode single-cell data consistently yields lower performance across multiple metrics. This finding suggests that the Q-Former’s specialized architecture is essential for capturing and processing the complex patterns inherent in single-cell gene expression, highlighting its critical role in our system.

Furthermore, since the Q-Former’s query embeddings aggregate information via attention mechanisms, we explore how varying the number of queries (1, 2, 4, 8, and 16) affects performance. Fig.~\ref{fig:closer}(c) shows that using 8 queries yields the best results, while increasing this number further leads to diminished performance. More queries provide the model with diverse, independent information streams, enabling each query to capture distinct aspects or features of the input single-cell data. However, an excessively large number of queries may introduce redundancy, making it harder for the model to distinguish relevant signals from noise and ultimately degrading performance. Thus, balancing query count is key to achieving both robust information coverage and optimal model accuracy.

\subsection*{Multi-task instruction tuning supports comprehensive understanding across tasks}

{\ours} is designed to handle multiple tasks within single-cell analysis, addressing the diverse demands of this field. Compared to single-task tuning, multi-task instruction tuning exposes the model to a broader range of problems during training, potentially fostering more general and adaptable representations.

To directly compare the performance gains from multi-task versus single-task training, we partition our multi-modal instruction dataset by task type and train separate single-task models for each. As illustrated in Fig.\ref{fig:closer} (d), we compute the average performance across all datasets for \textit{CTA} and \textit{DSP}. The results demonstrate that multi-task instruction tuning consistently outperforms its single-task counterpart on these tasks. This advantage likely arises from multi-task learning’s capacity to share parameters within the model architecture, thereby promoting knowledge transfer and integration across diverse tasks~\cite{DBLP:conf/pakdd/AwalCLM21,DBLP:journals/tkde/ZhangY22}. Moreover, balancing training across multiple tasks helps prevent overfitting to any single task, yielding a more well-rounded improvement in overall model performance.

For the \textit{CPCG} task, we use pKNN and {$\triangle$}sKNN as evaluation metrics, where a higher pKNN and a lower {$\triangle$}sKNN indicate superior performance. Under these criteria, multi-task instruction tuning also exhibits a clear advantage over single-task tuning. This suggests that even in complex generative scenarios requiring an understanding of underlying data distributions, leveraging knowledge across multiple tasks enhances the model’s capacity to capture detailed patterns, leading to more accurate and biologically meaningful pseudo-cell generation.

\subsection*{Pre-trained LM weights enhance classification performance of {\ours}}

To examine the influence of pre-trained weights on the performance of LMs in specialized single-cell analysis tasks, we alter the backbone of our {\ours}-\textit{chat} model by replacing the pre-trained LM with one initialized from scratch. As shown in Fig.~\ref{fig:closer}(e), the absence of pre-trained weights leads to marked performance declines in two classification tasks—\textit{CTA} and \textit{DSP}. However, for the \textit{CPCG} task, performance remains comparable, and the model without pre-trained LM weights even achieves a slight improvement on the $\triangle$sKNN metric.

This discrepancy arises because classification tasks heavily depend on the LM’s language processing and generation capabilities, which benefit significantly from exposure to diverse linguistic patterns and vocabularies during pre-training. In contrast, \textit{CPCG} relies more on dedicated downstream modules that reconstruct cells from hidden state vectors, making it less sensitive to the presence of pre-trained weights. This finding suggests that the effectiveness of the generation task is driven more by the specialized reconstruction module than by the LM’s pre-trained parameters.

\subsection*{Diversified instruction templates improve the validity and quality of {\ours} outputs}

To investigate how the number of instruction templates influences model performance, we evaluate {\ours} on previously unseen templates. As shown in Fig.~\ref{fig:closer}(f), we vary the ratio of available templates (0.5\%, 5\%, 50\%, and 100\% of the total) and assess the resulting performance.

Experiments on the \textit{CTA} and \textit{DSP} tasks indicate that increasing the number of templates leads to more robust performance, as reflected by reduced standard deviations. Although the results at 50\% and 100\% of the templates are comparable in terms of accuracy, the model trained with 100\% of the templates exhibits greater robustness. The exposure to more diverse instructions enhances the model’s ability to align its outputs more closely with the gene expression data. 
We do not consider the generation task in this analysis because the model sometimes fails to produce cells under the extremely low template ratios ($\leq$5\%) complicating the evaluation.

In addition, we use LLM-based assessments (Claude 3.5 Sonnet) to gauge response quality. The findings show that as the diversity of instruction templates increases, the model’s expressiveness improves, and performance variability decreases. This suggests that the model becomes more stable when trained on a broader range of instructions, a critical factor for practical applications in scientific research, where questions and data can be highly varied.

Lastly, we measure lexical diversity by calculating the unigram ratio, defined as the ratio of unique unigrams to the total number of unigrams in the response. The results indicate that as the variety of instruction templates grows, the lexical richness of the model’s output increases. This expanded vocabulary usage reflects the model’s capacity to produce more natural and contextually appropriate responses.

\section*{Discussion}

In this study, we present {\ours}, a multi-modal AI copilot that bridges the ``language of cellular biology'' and human natural language, establishing a foundation for advancing single-cell analysis. By developing a multi-modal single-cell instruction dataset spanning diverse species, tissues, and experimental conditions, we unify single-cell generation and understanding tasks within a cohesive framework. Building upon this dataset, {\ours} employs a multi-modal architecture capable of processing interleaved biological and textual data. By providing accurate, biologically relevant outputs tailored to researchers’ specific analytical needs, {\ours} lowers technical barriers and facilitates intuitive interaction with single-cell data through natural language. Extensive experiments reveal that single-cell generation and understanding tasks are mutually reinforcing, further enhancing the robustness and versatility of {\ours}.

Despite these advancements, {\ours} leaves room for future exploration and development. Expanding task coverage to include areas like predicting transcriptional responses to genetic perturbations~\cite{roohani2024predicting} or generating descriptive summaries of individual cells~\cite{levine2023cell2sentence} could significantly increase its versatility. Large-scale multitask instruction tuning~\cite{DBLP:conf/iclr/SanhWRBSACSRDBX22, wei2021finetuned, chung2024scaling, he2024zero} offers a promising avenue for improving adaptability and achieving zero-shot capabilities by leveraging extensive scRNA-seq datasets. Developing multi-round dialogue frameworks could enable richer and more interactive analyses. Furthermore, incorporating additional modalities beyond text and scRNA-seq—such as single-cell ATAC-seq~\cite{ashuach2022peakvi} or graph-based representations~\cite{liu2024muse}—would expand the scope of multi-modal single-cell analysis. These directions could pave the way for a more capable and generalizable AI copilot, driving forward single-cell research.

\section*{Methods}

\section*{Multi-modal instruction data construction}

\begin{figure*}[!th] 
  \centering
  \includegraphics[width=0.92\linewidth]{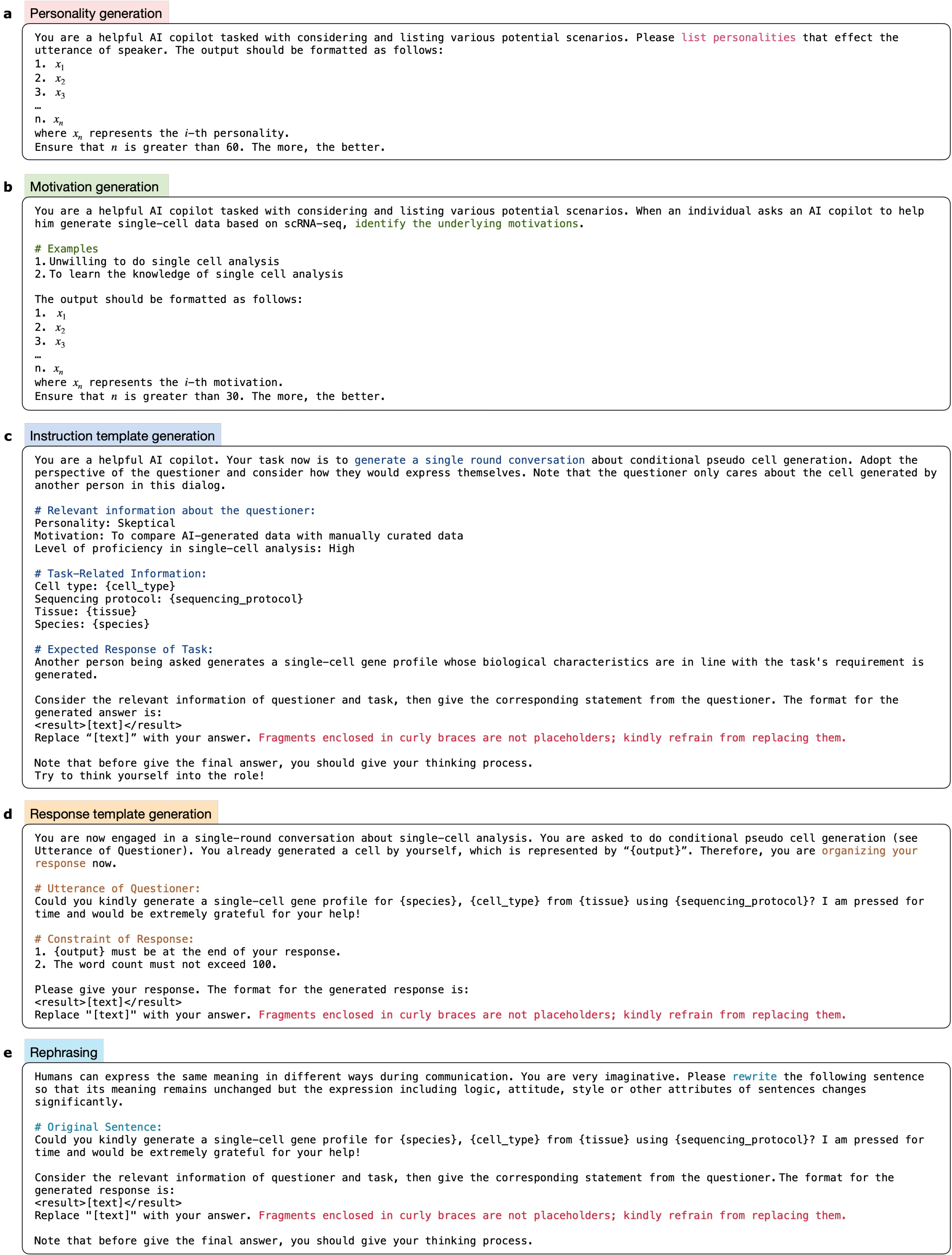}
  \caption{
  \small
  \textbf{The examples of the prompts used to construct instruction-response templates for \textit{CPCG}.}
  \textbf{a,} An example of the prompts for generating personality traits. \textbf{b,} An example of the prompts used to generate motivations for \textit{CPCG}. \textbf{c,} An example of the prompts used to generate instruction templates for \textit{CPCG}. \textbf{d,} An example of the prompts used to generate response templates for \textit{CPCG}. \textbf{e,} An example of the prompts for rewriting instruction templates to enhance diversity.
  }
  \vspace{-0.5cm}
  \label{fig:prompt}
\end{figure*}

We first gather publicly available scRNA-seq datasets for each task and convert them into multimodal instruction-following data using instruction‐response templates. However, constructing instruction-response templates for each task by hand is labor-intensive, resulting in a limited number of templates that often lack diversity. Such constraints can lead to trained models that are less robust and produce outputs with limited variety (Fig.~\ref{fig:closer}(f)). To address this issue, we propose leveraging an existing LLM to synthesize a wide range of instruction-response templates. We further increase template diversity by considering both the conversational context (i.e., the questioner’s motivation) and style (i.e., the questioner’s personality and level of expertise in single-cell analysis).

As illustrated in Fig.~\ref{fig:prompt}(a–b), we use GPT-4o~\cite{gpt4o} to generate various personality traits and each task's motivations. Experts then evaluate each motivation to ensure its validity. In contrast, personality traits are not reviewed, as GPT-4o reliably maps them to distinct conversational styles, even if a trait seems unrelated to the task. For example, the `generous’ trait often results in a more polite tone in the instruction templates. Additionally, we define two levels of proficiency in single-cell analysis, distinguishing between high and low expertise.

Next, we prompt GPT-4o to generate a set number of instruction-response templates for each task. During template generation, we specify the questioner’s personality trait, motivation, and familiarity level, asking GPT-4o to adopt the questioner’s perspective in formulating each instruction. We also provide GPT-4o with relevant placeholders (Fig.~\ref{fig:overview}(c)), emphasizing that these placeholders must appear in any generated instruction template. To help GPT-4o produce appropriate and coherent templates, we inform it of the expected response for each task. Fig.~\ref{fig:prompt}(c) demonstrates an example of the prompts used to synthesize instruction templates for \textit{CPCG}. Each generated template is filtered out if it exceeds 70 words or omits any essential placeholder. If it passes this filter, we calculate its ROUGE-L~\cite{lin2004rouge} similarity with previously generated instruction templates; if the maximum similarity surpasses 0.75, we ask GPT-4o to rewrite the instruction template (Fig.~\ref{fig:prompt}(e)). After three rewrites, if the highest similarity still exceeds 0.75, the instruction template is discarded.

We use GPT-4o to generate corresponding response templates for the instruction templates that pass the filter. Each response template includes the  \texttt{\{output\}} placeholder (Fig.~\ref{fig:prompt}(d)). In \textit{CTA}, \texttt{\{output\}} represents the cell type; in \textit{DSP}, it specifies how a single cell responds to a particular drug; and in \textit{CPCG}, it denotes the generated gene expression profile. For \textit{CPCG}, the response template must end with \texttt{\{output\}} and must not include ambiguous entities such as ``GENE A'' or ``Gene 1''. Response templates that fail to meet these criteria or exceed 70 words are filtered out, and the process proceeds to the next round of instruction-response template synthesis. Otherwise, we conclude that the instruction-response template has been successfully synthesized, and therefore retain it.

\section*{{\ours} model architecture}

\subsection*{Input embeddings}

{\ours} can take a mixture of text and single cells as input. Consider a sequence $\boldsymbol{X} = (\boldsymbol{x}_1, \boldsymbol{x}_2, \cdots, \boldsymbol{x}_m)$ of length $m$, where text and single cells are interleaved. Each single cell is encapsulated between two special tokens. Specifically, if $\boldsymbol{x}_i$ represents a single cell's gene expression profile, then $\boldsymbol{x}_{i-1}$ and $\boldsymbol{x}_{i+1}$ are marked with \textless CELL\textgreater\ and \textless /CELL\textgreater, respectively.

If $\boldsymbol{x}$ represents a text token within the input sequence, it is converted into a $d$-dimensional text embedding $\boldsymbol{e}$ by the LM's input embedding layer $\mathrm{embed}(\cdot; \boldsymbol{W})$:
\begin{equation}
\begin{aligned}
\boldsymbol{e} = \mathrm{embed}(\boldsymbol{x}; \boldsymbol{W}).
\end{aligned}
\end{equation}
Here, $\boldsymbol{W}$ is a learnable matrix with dimensions $(|V| + 3) \times d$, where $V$ is the original vocabulary of the pre-trained LM. Note that three special tokens,  \textless CELL\textgreater, \textless /CELL\textgreater, and \textless SIGNAL\textgreater, are added to the vocabulary. These tokens are initialized randomly at the beginning of training. 
On the other hand, if $\boldsymbol{x}$ is a single cell's gene expression profile in the input sequence, it is transformed into $k$ $(k \ge 1)$ cell embeddings of dimension $d$ by the cell encoder $\mathrm{qformer}(\cdot;\boldsymbol{\phi})$: 
\begin{equation}
\begin{aligned}
(\boldsymbol{e}_1, \boldsymbol{e}_2, \cdots, \boldsymbol{e}_k) = \mathrm{qformer}(\boldsymbol{x};\boldsymbol{\phi}),
\end{aligned}
\end{equation}
where $\boldsymbol{\phi}$ represents the learnable parameters of the cell encoder.

All input embeddings, whether derived from text tokens or single cells, are then concatenated in sequence to form the matrix $\boldsymbol{E}_{\boldsymbol{X}} = (\boldsymbol{e}_1, \boldsymbol{e}_2, \cdots, \boldsymbol{e}_n)$, where the $i$-th embedding $\boldsymbol{e}_i$ could be a text embedding or a cell embedding. 
This entire process is denoted as $\boldsymbol{E}_{\boldsymbol{X}} = \mathrm{map}(\boldsymbol{X}; \boldsymbol{W}, \boldsymbol{\phi})$.

\subsection*{Q-Former module}

The Q-Former module comprises three primary components: a set of learnable $d$-dimensional query vectors $\boldsymbol{Q}^{(0)} = (\boldsymbol{q}_1, \boldsymbol{q}_2, \cdots, \boldsymbol{q}_{k})$, a multi-layer perceptron (MLP) that encodes raw input cells into keys and values, and a transformer model~\cite{vaswani2017attention} that connects the outputs of the MLP to the query vectors.

Given an input cell $\boldsymbol{x}$, the MLP with residual connections~\cite{he2016deep} outputs a $t \times d$-dimensional vector $\boldsymbol{x}^{\mathrm{out}}$ which is then split into $t$ individual $d$-dimensional vectors:
\begin{equation}
\begin{aligned}
\boldsymbol{x}^{\mathrm{out}} &= \mathrm{MLP}(\boldsymbol{x}), \\
(\boldsymbol{u}_1, \boldsymbol{u}_2, \cdots, \boldsymbol{u}_t) &= \mathrm{split}(\boldsymbol{x}^{\mathrm{out}}, t),
\end{aligned}
\end{equation}
where $\mathrm{split}(\boldsymbol{x}^{\mathrm{out}}, t)$ denotes the operation of evenly dividing $\boldsymbol{x}^{\mathrm{out}}$ into $t$ vectors $\{\boldsymbol{u}_i\}_{i = 1}^{t}$. 
These vectors, denoted as $\boldsymbol{U} = (\boldsymbol{u}_1, \boldsymbol{u}_2, \cdots, \boldsymbol{u}_t)$, act as the keys and values for the transformer's cross-attention mechanism. Within each transformer block, the query vectors first interact with each other through a self-attention layer and then interact with $\boldsymbol{U}$ through a cross-attention layer~\cite{li2023blip}. This process is iteratively applied across $L_{\mathrm{cell}}$ stacked transformer blocks:
\begin{equation}
\begin{aligned}
\boldsymbol{Q}^{(l + 1)} = (\boldsymbol{q}_1^{(l + 1)}, \boldsymbol{q}_2^{(l + 1)}, \cdots, \boldsymbol{q}_{k}^{(l + 1)}) = \mathrm{qformerblock}^{(l)}(\boldsymbol{Q}^{(l)}, \boldsymbol{U}),
\end{aligned}
\end{equation}
where $0 \leq l < L_{\mathrm{cell}}$. The outcome of the entire computation process is represented as:
\begin{equation}
\begin{aligned}
(\boldsymbol{e}_1, \boldsymbol{e}_2, \cdots, \boldsymbol{e}_k) = \boldsymbol{Q}^{(L_{\mathrm{cell}})} = \mathrm{qformer}(\boldsymbol{x};\boldsymbol{\phi}).
\end{aligned}
\end{equation}

\subsection*{Cell reconstruction module}

The cell reconstruction module of {\ours} is a conditional variational autoencoder (CVAE) parameterized by $\boldsymbol{\psi} = (\boldsymbol{\psi}_e, \boldsymbol{\psi}_d)$ where $\boldsymbol{\psi}_e$ and $\boldsymbol{\psi}_d$ represent the encoder and decoder parameters, respectively.

During training, the encoder encodes the current condition $c$ and the corresponding single-cell $\boldsymbol{s}$ to produce a posterior distribution. The decoder then reconstructs the original input single-cell $\boldsymbol{s}$ from the latent variable $\boldsymbol{z}$ sampled from the posterior distribution $q(\boldsymbol{z}|\boldsymbol{s}, c; \boldsymbol{\psi}_e)$. The log probability density of the conditional distribution for the single-cell $\boldsymbol{s}$ can be expressed as:
\begin{equation}
\begin{aligned}
\log p(\boldsymbol{s}|c) &= \log \int p(\boldsymbol{s}, \boldsymbol{z}|c) \, d\boldsymbol{z} \\
&= \log \int \frac{p(\boldsymbol{s}, \boldsymbol{z}|c) q(\boldsymbol{z}|\boldsymbol{s}, c)}{q(\boldsymbol{z}|\boldsymbol{s}, c)} \, d\boldsymbol{z} \\
&\ge \mathbb{E}_{\boldsymbol{z} \sim q(\boldsymbol{z}|\boldsymbol{s}, c)} \left[ \log \frac{p(\boldsymbol{s}, \boldsymbol{z}|c)}{q(\boldsymbol{z}|\boldsymbol{s}, c)} \right] \\
&= \mathbb{E}_{\boldsymbol{z} \sim q(\boldsymbol{z}|\boldsymbol{s}, c)} \left[ \log p(\boldsymbol{s}|\boldsymbol{z}, c) \right] + \mathbb{E}_{\boldsymbol{z} \sim q(\boldsymbol{z}|\boldsymbol{s}, c)} \left[ \log \frac{p(\boldsymbol{z}|c)}{q(\boldsymbol{z}|\boldsymbol{s}, c)} \right] \\
&= \mathbb{E}_{\boldsymbol{z} \sim q(\boldsymbol{z}|\boldsymbol{s}, c; \boldsymbol{\psi}_e)} \left[ \log p(\boldsymbol{s}|\boldsymbol{z}, c; \boldsymbol{\psi}_d) \right] - \mathrm{KL}(q(\boldsymbol{z}|\boldsymbol{s}, c; \boldsymbol{\psi}_e)||p(\boldsymbol{z}|c)).
\end{aligned}
\end{equation}
The right-hand side of the inequality represents the evidence lower bound (ELBO). The training objective of the cell reconstruction module is to maximize the ELBO:~\cite{kingma2013auto}:
\begin{equation}
\begin{aligned}
\mathcal{L}_{\mathrm{recon}}(\boldsymbol{\psi}; \boldsymbol{s}, c) &= -\mathbb{E}_{\boldsymbol{z} \sim q(\boldsymbol{z}|\boldsymbol{s}, c; \boldsymbol{\psi}_e)} \left[ \log p(\boldsymbol{s}|\boldsymbol{z}, c; \boldsymbol{\psi}_d) \right] + \mathrm{KL}(q(\boldsymbol{z}|\boldsymbol{s}, c; \boldsymbol{\psi}_e)||p(\boldsymbol{z}|c)).
\end{aligned}
\end{equation}
scRNA-seq data have several key characteristics including overdispersion~\cite{love2014moderated,choudhary2022comparison}, sparsity~\cite{ding2020systematic,bouland2023consequences}, and non-negative integer expression values for each gene. Considering these properties, specifying the decoder-generated distribution as a Gaussian distribution is inappropriate. Researchers commonly analyze scRNA-seq data using either the Negative Binomial (NB) distribution or the Zero-Inflated Negative Binomial (ZINB) distribution~\cite{vieth2017powsimr,eraslan2019single}. For the NB distribution, the variance increases with the mean, which aligns with the overdispersion observed in scRNA-seq data. The ZINB distribution builds on the NB distribution by accounting for non-biological factors contributing to excess zeros. Since the NB distribution is encompassed by the ZINB distribution, we select the ZINB distribution as the output distribution of the decoder.

The inference and generative processes in the cell reconstruction module are similar to those in scVI~\cite{lopez2018deep}. Specifically, the latent variable $\boldsymbol{z}$ is divided into two independent variables $\boldsymbol{z}_s$ and $\ell$. Therefore, the loss function can be rewritten as:
\begin{equation}
\begin{aligned}
\mathcal{L}_{\mathrm{recon}}(\boldsymbol{\psi}; \boldsymbol{s}, c) = &-\mathbb{E}_{\boldsymbol{z}_{\boldsymbol{s}} \sim q(\boldsymbol{z}_{\boldsymbol{s}}|\boldsymbol{s}, c; \boldsymbol{\psi}_e), \ell \sim q(\ell|\boldsymbol{s}, c; \boldsymbol{\psi}_e)} \left[ \log p(\boldsymbol{s}|\boldsymbol{z}_{\boldsymbol{s}}, \ell, c; \boldsymbol{\psi}_d) \right] \\
&+ \alpha \left[ \mathrm{KL}(q(\boldsymbol{z}_{\boldsymbol{s}}|\boldsymbol{s}, c; \boldsymbol{\psi}_e)||p(\boldsymbol{z}_{\boldsymbol{s}}|c)) + \mathrm{KL}(q(\ell|\boldsymbol{s}, c; \boldsymbol{\psi}_e)||p(\ell|c)) \right],
\end{aligned}
\end{equation}
where $\alpha$ balances the trade-off between the reconstruction quality and output diversity of CVAE~\cite{shao2020controlvae}.
We specify $p(\boldsymbol{z}_{\boldsymbol{s}}|c)$ as an isotropic Gaussian distribution $\mathcal{N}(\boldsymbol{0}, \boldsymbol{I})$, and $p(\ell|c)$ as a log-normal distribution $\mathrm{Lognormal}(\ell_{\mu}, \ell_{\sigma}^2)$ where $\ell_{\mu} = \mathbb{E}_{\boldsymbol{s}}[\log(\sum_{i} s_i)]$, and $\ell_{\sigma}^2 = \mathbb{E}_{\boldsymbol{s}}[(\log(\sum_{i} s_i) - \ell_{\mu})^2]$. 

To disentangle the sampling operation from the optimization, we employ the reparameterization trick. The encoder $\boldsymbol{\psi}_e$ outputs the means and variances of the distributions based on the input cell $\boldsymbol{s}$ and the condition $c$. We perform the sampling operation by drawing samples from a standard normal distribution $\mathcal{N}(\boldsymbol{0}, \boldsymbol{I})$. These samples are subsequently scaled by the computed variances and shifted by the computed means. The Kullback-Leibler divergence between two Gaussian distributions has a closed-form solution:
\begin{equation}
\begin{aligned}
\mathrm{KL}(\mathcal N(\boldsymbol{x} &;\boldsymbol{\mu}_1,\boldsymbol{\Sigma}_1) || \mathcal N(\boldsymbol{y};\boldsymbol{\mu}_2,\boldsymbol{\Sigma}_2)) \\
&= \frac{1}{2}[\log \frac{|\boldsymbol{\Sigma}_2|}{|\boldsymbol{\Sigma}_1|} - d + \mathrm{tr}({\boldsymbol{\Sigma}_2}^{-1} {\boldsymbol{\Sigma}_1}) + {(\boldsymbol{\mu}_2 - \boldsymbol{\mu}_1)}^{T} {\boldsymbol{\Sigma}_2}^{-1} (\boldsymbol{\mu}_2 - \boldsymbol{\mu}_1)], 
\end{aligned}
\end{equation}
where $d$ is the dimensionality of the random variables $\boldsymbol{x}$ and $\boldsymbol{y}$, and $\mathrm{tr}(\cdot)$ denotes the trace of a matrix. We can efficiently compute the second term and third term in the loss function using the means and covariances of the target and estimated distributions.

The generative process is defined as a sequence of transformation and sampling operations, described as follows:
\begin{equation}
\begin{aligned}
\boldsymbol{z}_{\boldsymbol{s}} &\sim \mathrm{Normal}(\boldsymbol{0}, \boldsymbol{I}), \\
\ell &\sim \mathrm{Lognormal}(\ell_{\mu}, \ell_{\sigma}^2), \\
\ell^{\prime} &= f_1(\ell, c), \\
\boldsymbol{\rho} &= f_2(\boldsymbol{z}_{\boldsymbol{s}}, c), \\
g_i &= f_3(i), \\
w_i &\sim \mathrm{Gamma}(\rho_i, g_i), \\
v_i &\sim \mathrm{Poisson}(\ell^{\prime} w_i), \\
\boldsymbol{\tau} &= f_4(\boldsymbol{z}_{\boldsymbol{s}}, c), \\
b_i &\sim \mathrm{Bernoulli}(\tau_i), \\
\hat{s}_i &= 
	\begin{cases}
		v_i, & b_i = 0 \\
		0, & b_i = 1
	\end{cases}.
\end{aligned}
\end{equation}
Here, $f_1$, $f_2$, $f_3$, and $f_4$ are the mapping functions parameterized by $\boldsymbol{\psi}_d$. $g_i$ represents the inverse dispersion of the $i$-th gene, and $\boldsymbol{\rho}$ ensures $\sum_{i} \rho_i = 1$. During training, $\boldsymbol{z}_{\boldsymbol{s}}$ and $\ell$ are sampled from their respective posterior distributions $q(\boldsymbol{z}_{\boldsymbol{s}}|\boldsymbol{s}, c; \boldsymbol{\psi}_e)$ and $q(\ell|\boldsymbol{s}, c; \boldsymbol{\psi}_e)$. The model then computes the log probability of $\boldsymbol{s}$, $\log p(\boldsymbol{s}|\boldsymbol{z}_{\boldsymbol{s}}, \ell, c; \boldsymbol{\psi}_d)$, under the generated ZINB distribution, which is a stable computational process~\cite{lopez2018deep}. Therefore, the first term of the loss function can also be efficiently computed. Note that if no conditional information is necessary, $c$ can be set to a null vector $\emptyset$~\cite{ho2022classifier}.

\subsection*{Pre-trained LM}

The architecture of pretrained language models (PLMs) can be categorized as encoder-only~\cite{devlin2018bert,liu2019roberta,lan2019albert,he2020deberta}, decoder-only~\cite{radford2018improving,brown2020language,wang2021gpt,DBLP:journals/corr/abs-2302-13971}, or encoder-decoder~\cite{DBLP:conf/nips/00040WWLWGZH19,DBLP:conf/acl/LewisLGGMLSZ20,raffel2020exploring,zhang2020pegasus}. Since the three single-cell tasks considered in this study can all be formulated as sequence-to-sequence (seq2seq) tasks, any decoder-only or encoder-decoder PLM can flexibly serve as the backbone network of {\ours}. To simplify the notation in mathematical formulations, we denote the backbone network's forward computation process (excluding the transformation of input sequences into input embeddings) as $p_{\mathrm{PLM}}(\cdot; \boldsymbol{\theta})$.

\paragraph{Single-cell understanding task}
For the single-cell understanding task, given an input sequence $\boldsymbol{X}$ and the corresponding target sequence $\boldsymbol{Y} = (\boldsymbol{y}_1, \boldsymbol{y}_2, \cdots, \boldsymbol{y}_{L_{\mathrm{target}}})$, where $\boldsymbol{y}_i$ is the $i$-th token in the target sequence, the objective of LM is to minimize the negative log joint probability of the target sequence~\cite{devlin2018bert}:
\begin{equation}
\begin{aligned}
\mathcal{L}_{\mathrm{scu}}(\boldsymbol{\theta}, \boldsymbol{W}, \boldsymbol{\phi}; \boldsymbol{X}, \boldsymbol{Y}) 
&= -\sum_{i=1}^{L_{\mathrm{target}}} \log p_{\mathrm{PLM}}(\boldsymbol{y}_{i}|\mathrm{map}(\boldsymbol{X}; \boldsymbol{W}, \boldsymbol{\phi}), \mathrm{map}(\boldsymbol{Y}_{i-1}; \boldsymbol{W}, \boldsymbol{\phi}); \boldsymbol{\theta}) \\
&= -\sum_{i=1}^{L_{\mathrm{target}}} \log p_{\mathrm{PLM}}(\boldsymbol{y}_{i}|\boldsymbol{E}_{\boldsymbol{X}}, \boldsymbol{E}_{\boldsymbol{Y}_{i-1}}; \boldsymbol{\theta}) \\
&= \mathcal{L}_{\mathrm{text}}(\boldsymbol{\theta}, \boldsymbol{W}, \boldsymbol{\phi}; \boldsymbol{X}, \boldsymbol{Y}),
\end{aligned}
\end{equation}
where $\boldsymbol{Y}_{i-1} = (\boldsymbol{y}_1, \boldsymbol{y}_2, \cdots, \boldsymbol{y}_{i-1})$.

\paragraph{Single-cell generation task}
In the single-cell generation task, the target sequence $\boldsymbol{Y}$, with a length of $L_{\mathrm{target}}$, ends with a special signal token \texttt{<SIGNAL>} that marks the transition to generating a single cell. In this task, the LM's objective is not only to generate the target sequence but also to produce conditional information $\boldsymbol{h}_{\texttt{<CPCG>}}$ based on $(\boldsymbol{X}, \boldsymbol{Y})$. This conditional information $\boldsymbol{h}_{\texttt{<CPCG>}}$ enables the decoder in the cell reconstruction module to generate the target single cell $\boldsymbol{s}$. The overall process can be formalized as follows:
\begin{equation}
\begin{aligned}
\mathcal{L}_{\mathrm{scg}}(\boldsymbol{\theta}, \boldsymbol{W}, \boldsymbol{\phi}, \boldsymbol{\psi}; \boldsymbol{X}, \boldsymbol{Y}, \boldsymbol{s}) &= \mathcal{L}_{\mathrm{text}}(\boldsymbol{\theta}, \boldsymbol{W}, \boldsymbol{\phi}; \boldsymbol{X}, \boldsymbol{Y}) + \mathcal{L}_{\mathrm{recon}}(\boldsymbol{\psi}; \boldsymbol{s}, \boldsymbol{h}_{\texttt{<CPCG>}}), 
\end{aligned}
\end{equation}
where, in this study, $\boldsymbol{h}_{\texttt{<CPCG>}}$ represents the hidden state of the \texttt{<SIGNAL>} token in the last layer of the LM. By leveraging the LM as an encoder for the biological characteristics of single cells, {\ours} enhances the flexibility and expressiveness of user-specified input conditions.

Collectively, {\ours} can be trained end-to-end and the corresponding optimization objective is defined as:
\begin{equation}
\begin{aligned}
\underset {\boldsymbol{\theta}, \boldsymbol{W}, \boldsymbol{\phi}, \boldsymbol{\psi}}{\operatorname {arg\,min}}\; \mathbb{E}_{(\boldsymbol{X}, \boldsymbol{Y}, \boldsymbol{s}) \sim \mathcal{D}} [\mathbb{I}(\boldsymbol{s} = \emptyset) \mathcal{L}_{\mathrm{scu}}(\boldsymbol{\theta}, \boldsymbol{W}, \boldsymbol{\phi}; \boldsymbol{X}, \boldsymbol{Y}) + \mathbb{I}(\boldsymbol{s} \neq \emptyset) \mathcal{L}_{\mathrm{scg}}(\boldsymbol{\theta}, \boldsymbol{W}, \boldsymbol{\phi}, \boldsymbol{\psi}; \boldsymbol{X}, \boldsymbol{Y}, \boldsymbol{s})],
\end{aligned}
\end{equation}
where $\mathcal{D}$ represents the training set, $\boldsymbol{s} = \emptyset$ indicates the absence of the cell modality in the output, and $\mathbb{I}(\cdot)$ is the indicator function. From the perspective of multitask learning, the majority of the model's parameters (i.e., the LM without the language modeling head) are shared across multiple tasks, which effectively prevents overfitting on any single task~\cite{baxter1997bayesian,ruder2017overview}.

\section*{Evaluation}

\paragraph{Answer extraction}

Unlike {\ours}-\textit{instruct}, {\ours}-\textit{chat} generates free-form responses, making direct evaluation of its performance challenging. For the \textit{CPCG} task, the generated gene expression profiles are placed at the end of the output, which enables straightforward extraction and analysis. In contrast, answers for the other two classification tasks are embedded within conversational text, which requires a reliable extraction tool. To address this, we employ xFinder~\cite{DBLP:journals/corr/abs-2405-11874}, a state-of-the-art extraction model whose reliability is evaluated meticulously in our setting. 
Specifically, we first construct a test set by replacing the \texttt{\{output\}} placeholders in single‐round dialogue templates with the labels derived from all datasets. For example, GSE110894 contains 2 distinct labels (Figure~\ref{fig:data_distribution}). Since there are 2,359 different templates used for \textit{DSP} (see the section Instruction‐response template construction details), we can construct 4,718 test samples.  Overall, the test set comprises 178,411 samples. We find that xFinder achieves an accuracy exceeding 99.97\% on this test set, demonstrating its effectiveness in automatically extracting answers.

\paragraph{Metrics}

For the \textit{CTA} and \textit{DSP} tasks, we use standard evaluation metrics, including weighted F1, macro F1, and accuracy. 
When evaluating {\ours}-\textit{chat}, there are rare instances where the model fails to provide an answer. In calculating accuracy, we use the `true accuracy' metric, treating unanswered cases as incorrect predictions. However, for F1 scores, we exclude unanswered cases because F1 relies on valid predictions for precision and recall calculations. Including unanswered cases as errors would misrepresent the balance between precision and recall, leading to an inaccurate evaluation of the model’s performance.

For the \textit{CPCG} task, we evaluate the model’s performance using Maximum Mean Discrepancy (MMD), sKNN, and pKNN.
Specifically, for a real single-cell dataset $\{(\boldsymbol{y}_i, c_i)\}_{i = 1}^M$ unseen during the training, the trained model generates a pseudo single-cell dataset $\{(\boldsymbol{x}_i, c_i)\}_{i = 1}^M$ based on the corresponding label set $\{c_i\}_{i = 1}^M$.
Since both real and generated single-cell data are count data, we normalize each sample to ensure the gene counts sum to 10,000, followed by log1p-transformation.
To evaluate the model, we first apply principal component analysis (PCA) to reduce the dimensionality of the real single-cell data to 50 dimensions, then further reduce it to a two-dimensional embedding space using UMAP~\cite{mcinnes2018umap}. The pseudo single-cell data is mapped into the same two-dimensional embedding space. Finally, we compute MMD, sKNN, and pKNN within this embedding space to assess the model’s performance.

\textbf{MMD} is a kernel-based statistical test, used to measure the difference between two probability distributions~\cite{gretton2012kernel}. 
Given samples $\{\boldsymbol{x}_i\}_{i=1}^M$ from the estimated distribution $p$ and $\{\boldsymbol{y}_j\}_{j=1}^M$ from the real distribution $q$, the empirical estimate of MMD between $p$ and $q$ can be calculated using the following formula:
\begin{equation}
\begin{aligned}
\operatorname{MMD}[p, q] = \left\| \frac{1}{M} \sum_{i=1}^M \phi(\boldsymbol{x}_i) - \frac{1}{M} \sum_{j=1}^M \phi(\boldsymbol{y}_j) \right\|_{\mathcal{H}},
\end{aligned}
\end{equation}
where $\mathcal{H}$ is a universal Reproducing Kernel Hilbert Space (RKHS) and $\phi(\cdot)$ is the corresponding reproducing kernel. Similar to \cite{shaham2017removal, marouf2020realistic}, we use a kernel function $\mathcal{K}(\boldsymbol{x}, \boldsymbol{y})$ that is the summation of three different Gaussian kernels to compute the MMD:
\begin{equation}
\begin{aligned}
\mathcal{K}(\boldsymbol{x}, \boldsymbol{y}) &= \sum_{i=1}^{3} \exp(-\gamma_{i} \|\boldsymbol{x} - \boldsymbol{y}\|^2), \\
\gamma_{i} &= \frac{1}{2^{i - 2} \omega^2},  \\
\operatorname{MMD}[p, q] &= \sqrt{\frac{1}{M^2} \sum_{i,j=1}^M \mathcal{K}(\boldsymbol{x}_i, \boldsymbol{x}_j) - \frac{2}{M^2} \sum_{i=1}^M \sum_{j=1}^M \mathcal{K}(\boldsymbol{x}_i, \boldsymbol{y}_j) + \frac{1}{M^2} \sum_{i,j=1}^M \mathcal{K}(\boldsymbol{y}_i, \boldsymbol{y}_j)},
\end{aligned}
\end{equation}
where $\omega$ is the median of the average distance between a real single cell to its nearest 25 neighbors. A smaller MMD value indicates that the learned distribution $p$ is more similar to the target distribution $q$.

\textbf{sKNN} focuses on evaluating the intrinsic biological structure of single-cell data. It is calculated by finding the $K$ nearest neighbors for each cell, and then evaluating the label consistency based on the neighbors’ labels:
\begin{equation}
\begin{aligned}
\operatorname{sKNN}=\frac{1}{M} \sum_{i=1}^M \frac{1}{K} \sum_{\substack{j \in \text{Neighbor}_{K}(\boldsymbol{x}_i)}} \mathbb{I}\left(c_j=c_i\right),
\end{aligned}
\end{equation}
where $\text{Neighbor}_{K}(\boldsymbol{x}_i)$ denotes the $K$ nearest neighbors of the $i$-th cell $\boldsymbol{x}_i$. The function $\mathbb{I}(\cdot)$ is an indicator function that returns 1 if the condition is true (i.e., if two labels match) and 0 otherwise. In our evaluation setting, a label corresponds to a cell type. It is important to note that a low sKNN value does not necessarily indicate a significant discrepancy between the distribution learned by the model and the real distribution. This may occur if the clusters formed by each cell type in the real single-cell dataset are not compact. Therefore, it is recommended to calculate sKNN on the real single-cell dataset as a reference. 
In our evaluation setting, we report $\triangle$sKNN, which represents the absolute difference between the sKNN values of the generated data and the real data:
\begin{equation}
\begin{aligned}
\triangle\operatorname{sKNN} = \left| \operatorname{sKNN}_{\text{generated}} - \operatorname{sKNN}_{\text{real}} \right|.
\end{aligned}
\end{equation}
A smaller $\triangle$sKNN value indicates a closer alignment of the intrinsic biological structure between the generated and real data, while a larger value implies greater divergence.
 
\textbf{pKNN}, in contrast, evaluates whether the label of each pseudo cell aligns with the majority label of its $K$ nearest real cells, thereby assessing the positional proximity of the generated cells to the real cells. To compute pKNN, we use $\{(\boldsymbol{y}_i, c_i)\}_{i = 1}^M$, the real single-cell data with their corresponding labels, as the training set for a KNN classifier, and $\{(\boldsymbol{x}_i, c_i)\}_{i = 1}^M$, the model-generated single-cell data with their corresponding labels, as the test set. The pKNN metric is then defined as the classification accuracy of this KNN classifier when predicting the labels of the generated cells. A higher pKNN value indicates a closer alignment of the spatial distribution between the generated and real data.

Overall, these metrics form a holistic evaluation framework spanning all three tasks. For the two classification tasks, macro F1 accounts for class imbalances, while weighted F1 emphasizes the model's performance on majority classes. For \textit{CPCG}, the two KNN-based metrics focus on different aspects of local patterns, whereas MMD captures global distributional differences.

\paragraph{Extraction of key genes using gradient saliency-based methods}

We use the Vanilla Gradient method~\cite{DBLP:journals/corr/SimonyanVZ13} to identify genes that play a significant role in the model’s decision-making process. Although this method is relatively simple to implement, it is effective for extracting important features because it is sensitive to both the model’s parameters and the dataset~\cite{DBLP:conf/nips/AdebayoGMGHK18}.
Given an input sequence from the test set, denoted as $\boldsymbol{X} = (\boldsymbol{x}_1, \boldsymbol{x}_2, \cdots, \boldsymbol{x}_m)$, where the $i$-th element $\boldsymbol{x}_i$ represents a single cell's gene expression profile $\boldsymbol{s}$, we compute the gradient $\boldsymbol{g}_{\boldsymbol{s}}$ as follows:
\begin{equation}
\begin{aligned}
\boldsymbol{g}_{\boldsymbol{s}} = \frac{\partial \mathcal{L}}{\partial \boldsymbol{x}_i} \bigg|_{\boldsymbol{x}_i= \boldsymbol{s}},
\end{aligned}
\end{equation}
where $\mathcal{L}$ is the loss function with respect to $\boldsymbol{x}_i$, as defined by the trained model and the corresponding target sequence $\boldsymbol{Y}$. We define $\mathcal{G}$ as the set of indices that correspond to the genes of interest, such as all protein-coding genes in the gene vocabulary. We then compute a mask $\boldsymbol{m}_{\boldsymbol{s}}$, where the $i$-th element $m_{s_i}$ is defined as:
\begin{equation}
\begin{aligned}
m_{s_i} = \begin{cases} 
1, & \text{if } i \in \mathcal{G} \land g_{s_i} < 0 \land s_i > 0, \\
0, & \text{otherwise}. 
\end{cases}
\end{aligned}
\end{equation}
Here, $g_{s_i}$ represents the $i$-th element of $\boldsymbol{g}_{\boldsymbol{s}}$. We subsequently derive the feature importance scores $\boldsymbol{o}_{\boldsymbol{s}}$ as follows:
\begin{equation}
\begin{aligned}
\boldsymbol{o}_{\boldsymbol{s}} = - \boldsymbol{g}_{\boldsymbol{s}} \otimes \boldsymbol{m}_{\boldsymbol{s}}.
\end{aligned}
\end{equation}
We apply min-max normalization to the elements of $\boldsymbol{o}_{\boldsymbol{s}}$, scaling them to the range [0, 1]. After normalization, we sort the elements of $\boldsymbol{o}_{\boldsymbol{s}}$ in descending order, considering the top-ranked gene as the most critical for determining the cell type of that sample. Given that the test set typically contains multiple single cells, we compute the normalized feature importance scores for each cell. We then group the feature importance scores by cell type and perform an average aggregation, deriving aggregated feature importance scores for each cell type. Ultimately, we identify the top $n$ (where $n=10$) significant genes for each cell type, treating them as potential marker genes.

\paragraph{UMAP visualizations}

Dimensionality reduction methods such as PCA, t-SNE~\cite{van2008visualizing}, and UMAP~\cite{mcinnes2018umap} are widely used to visualize high-dimensional data. In this study, we select UMAP because of its effectiveness in visualizing scRNA-seq data~\cite{becht2019dimensionality}. For an scRNA-seq dataset to be visualized, we first normalize each sample to ensure that the gene counts sum to 10,000. Following normalization, log1p-transformation is applied to each sample. Next, we compute the top 50 principal components using the PCA algorithm and further reduce the dimensionality to two dimensions with UMAP. Finally, the resulting two-dimensional data points are visualized.

\subsection*{Experimental setup}

\paragraph{Datasets and gene vocabulary}

All datasets used in this study are publicly available. We collect all raw cell-by-gene matrices and employ Scanpy toolkit~\cite{wolf2018scanpy} to preprocess them independently. Specifically, for the purpose of quality control~\cite{mccarthy2017scater}, we filtered out single cells where fewer than 200 unique genes are expressed.  After removing low-quality cells, we exclude genes expressed in fewer than 8 different single cells. Aberrant cells, such as those with the excessive number of mitochondrial genes expressed or abnormally high total counts, are also removed. Since the original data from Xin-2016~\cite{xin2016rna} does not provide cell type labels, we annotate each sample using cell-type-specific marker genes based on the findings of Muraro et al.~\cite{muraro2016single} and the method employed by Tritschler et al.~\cite{tritschler2022transcriptional}. For all datasets used for \textit{CTA}, rare cell types (i.e., those with fewer than 20 cells) are excluded.

To harmonize gene features across two species, we process the datasets collected from mice. Concretely, we convert provided gene symbols into Ensembl gene IDs using pyEnsembl~\cite{pyEnsembl}. We then utilize pyBiomart~\cite{pyBiomart} to map mouse genes to their human orthologs, whereas mouse-specific genes are kept. To standardize gene features across all datasets, we identify the top 3,600 highly variable genes for each dataset's cell-by-gene matrix using Seurat~\cite{satija2015spatial, stuart2019comprehensive}. These sets of highly variable genes are merged to create a unified gene vocabulary. The resulting gene vocabulary, containing 18,961 unique genes, is denoted as $V_{g}$. Finally, each dataset is divided into training, validation, and test sets in an 8:1:1 ratio.

\begin{figure}[!h]
\centering
\vspace{-0.5cm}
\begin{minipage}{\textwidth}
\centering
\scalebox{0.6}{
\begin{tabular}{c|c|c|c|c|c|c}
\toprule

\multicolumn{1}{c|}{{Task Type}} & \multicolumn{1}{c}{{Species}} & \multicolumn{1}{|c}{{Dataset}} & \multicolumn{1}{|c}{{Protocol}} & \multicolumn{1}{|c|}{{Drug}} & Tissue & \# Samples \\

\midrule

\multirow{5}{*}{Cell type annotation}  & \multirow{3}{*}{Human} & He-2020-Liver~\cite{he2020single}  & HiSeq X Ten System  & - & Liver & 2561 \\

 \cline{3-7}
 
 & & Segerstolpe-2016~\cite{segerstolpe2016single} & Smart-seq2 & - & Pancreas & 2188 \\

 \cline{3-7}
 
 & & Xin-2016~\cite{xin2016rna} & SMARTer & - & Pancreas & 20528 \\
 
\cline{2-7}  

& \multirow{2}{*}{Mouse} & Ma-2020~\cite{ma2020chromatin} & SHARE-seq & - & Skin & 34135 \\

\cline{3-7}

& & Bastidas-Ponce-2019~\cite{bastidas2019comprehensive} & 10x & - & Pancreas & 35481 \\

\midrule

\multirow{3}{*}{Drug sensitivity prediction} & \multirow{2}{*}{Human} & GSE117872~\cite{sharma2018longitudinal} & Fluidigm C1 & Cisplatin & Oral cavity & 1302 \\

\cline{3-7}

& & GSE149383~\cite{aissa2021single} & Drop-seq & Erlotinib & Lung & 2254 \\

\cline{2-7}

 & Mouse & GSE110894~\cite{bell2019targeting} & Cel-Seq2 & BET inhibitor (I-BET-762) & Bone marrow & 1419 \\

\midrule

\multirow{11}{*}{Conditional pseudo-cell generation} & \multirow{6}{*}{Human} & PBMC68K~\cite{zheng2017massively} & 10x & - & Blood & 68185 \\

\cline{3-7}

& & \multirow{5}{*}{Tabular-Sapiens~\cite{the2022tabula}} & \multirow{5}{*}{10x} & \multirow{5}{*}{-} & Spleen & 12864 \\

& & & & & Blood & 5129 \\

& & & & & Thymus & 7842 \\

& & & & & Vasculature & 8112 \\

& & & & & Bladder & 11404 \\

\cline{2-7}

& \multirow{5}{*}{Mouse} & \multirow{5}{*}{Mouse-Atlas~\cite{tabula2020single}} & \multirow{5}{*}{10x} & \multirow{5}{*}{-} & Liver & 7292 \\

& & & & & Spleen & 35718 \\

& & & & & Thymus & 9275 \\

& & & & & Bladder & 8945 \\

& & & & & Lung & 24521 \\

\bottomrule
\end{tabular}
}
\end{minipage}

\begin{minipage}{\textwidth}
\vspace{0.2cm}
    \centering
    \includegraphics[width=0.86\textwidth]{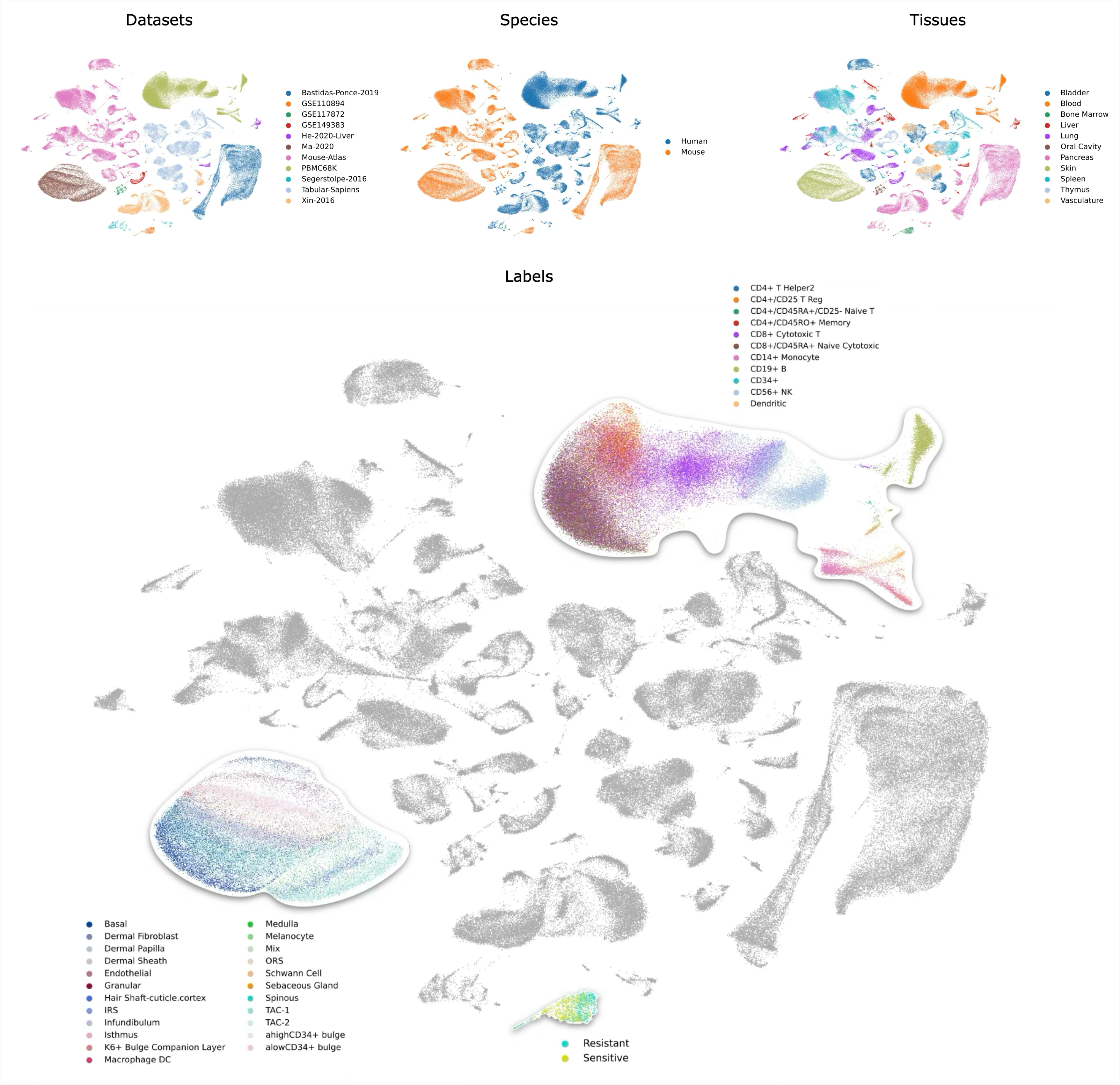} 
\end{minipage}
\caption{
\small
\textbf{Detailed overview and UMAP visualizations of scRNA-seq datasets used in this work.} 
A total of 11 datasets are utilized, spanning 2 species and 11 tissue types. Among them, 5 datasets are employed for \textit{CTA}, 3 for \textit{DSP}, and 3 for \textit{CPCG}. The number of samples for each dataset is listed under the column \# Samples. The middle three UMAP plots present the single-cell data landscape, showcasing the distributions across different datasets, species, and tissues. The lower UMAP plot shows the label distribution for Ma-2020, GSE110894, and PBMC68K. Covering three distinct tasks, our data include various labels, such as response labels for \textit{DSP} and cell type labels for other two tasks.
}
\label{fig:data_distribution}
\end{figure}

Figure~\ref{fig:data_distribution} provides a detailed overview and UMAP visualizations of the processed datasets. These datasets originate from diverse sources encompassing two species and eleven distinct tissue types. The datasets also exhibit significant variation in complexity. For the \textit{CPCG} task, Mouse-Atlas contains 50 distinct cell types, whereas Tabular-Sapiens contains only 6 distinct cell types. For the \textit{CTA} task, as illustrated in Figure~\ref{fig:cell_type}(a), the performance of baselines and {\ours} varies substantially across Ma-2020 and Segerstolpe-2016. Collectively, these datasets enable a comprehensive evaluation of {\ours} across the three tasks.

\paragraph{Instruction-response template construction details}

The construction of instruction-response templates is based on the GPT-4o model~\cite{gpt4o}. The hyperparameter of decoding algorithm, $top\_p$, is set to 0.95. Initially, we leverage the world knowledge of GPT-4o to generate a list of relevant personality traits and each task's motivations. All 79 personality traits listed by GPT-4o are considered, and the listed motivations for each task are manually checked.  After the review,  the numbers of motivations for \textit{CTA}, \textit{DSP}, and \textit{CPCG} are 40, 50, and 47, respectively. 

As discussed in the multi-modal instruction data construction section, three types of traits are considered during the synthesis of instruction templates: personality, motivation, and proficiency. To improve the diversity of instruction templates, randomness is introduced in the number of trait types included in the prompt used for the synthesis of instruction templates (Fig.~\ref{fig:prompt}(c)). Specifically, there is a 27\% probability of including all three types of traits, a 45\% probability of including two randomly selected traits, an 18\% probability of including one randomly selected trait, and a 10\% probability that no traits related to the questioner are provided.  

To accelerate the synthesis process, we generate instruction-response templates in parallel, using 10 different API keys with 10 distinct random seeds independently. Each API key is invoked 1,080 times (360 instruction-response pairs for each task). For two classification tasks (\textit{CTA} and \textit{DSP}), there is a 50\% probability that the prompt will require the model to generate an instruction template containing an option placeholder \texttt{\{option\}}. After synthesis, the resulting 10 sets of instruction-response templates are merged. We define the similarity between two instruction-response pairs as the value of ROUGE-L~\cite{lin2004rouge} between their instruction templates. Instruction-response pairs with a maximum similarity score greater than 0.75 are discarded to ensure diversity. The final numbers of instruction-response templates for \textit{CTA}, \textit{DSP}, and \textit{CPCG} are 2,787, 2,395, and 2,080, respectively. Lastly, each task’s instruction-response templates are split into training, validation, and test sets in an 8:1:1 ratio.

\paragraph{Implementation details}

{\ours} employs the T5-base~\cite{raffel2020exploring} model as the backbone. For the Q-Former module, $L_{\mathrm{cell}}$ is set to 4, the number of query tokens $k$ is 8, and the number of key-value pairs $t$ is 6. The MLP in the Q-Former module is composed of three fully connected layers, with skip connections present in the last two layers. A fully connected layer connects the cell reconstruction module to the language model, transforming the dimension of the hidden state $\boldsymbol{h}_{\mathrm{<CPCG>}}$ from 768 to 256. The resulting output serves as a conditioning vector for the cell reconstruction module. The cell reconstruction module consists of an encoder and a decoder. The encoder $\boldsymbol{\psi}_{e}$ contains two separate MLPs which compute posterior distributions for $\boldsymbol{z}_{\boldsymbol{s}}$ and $\ell$ respectively. The decoder $\boldsymbol{\psi}_{d}$ includes three distinct MLPs—$f_1$, $f_2$, and $f_4$—with $f_2$ and $f_4$ sharing parameters except in the final layer. A learnable vector $\boldsymbol{g}$, with a dimension equal to the size of gene vocabulary, parameterizes $f_3$. The dimension of the latent variable $\boldsymbol{z}_{\boldsymbol{s}}$ is set to 256.

All experiments related to {\ours} are conducted on a single-node server equipped with eight 32GB V100 GPUs. For multi-task instruction tuning, we use the gene vocabulary $V_{g}$ to unify gene features across all datasets. We then merge all training splits into a single mixed training set and format each sample using a template randomly selected from the constructed set of instruction-response templates. Likewise, this process is also applied to generate the mixed validation set and test set. We train {\ours} using the Adafactor optimizer~\cite{shazeer2018adafactor} at a learning rate of $1 \times 10^{-3}$ for a total of 160 epochs. The batch size is set to 32 for {\ours}-\textit{chat} and 64 for {\ours}-\textit{instruct}. In particular, all response templates used for training {\ours}-\textit{instruct} are replaced with \texttt{\{output\}}. The model with the lowest loss computed on the validation set is selected for evaluation.

\section*{Appendix}

\section*{Related work}

\subsection*{Single-cell analysis}

Single-cell analysis studies individual cells to understand their roles in biological systems. A key tool in this field is single-cell RNA sequencing (scRNA-seq), which measures the activity of genes in each cell and their expression levels~\cite{plass2018cell,cao2019single}. The data from scRNA-seq are arranged into gene expression matrices, where rows represent genes, columns represent cells, and the numbers show how much each gene is expressed~\cite{brazma2000gene}. These matrices allow researchers to investigate tasks like identifying different cell types, studying tissue composition, and exploring changes in cell states.
In addition to scRNA-seq, other technologies such as single-cell ATAC-seq and spatial transcriptomics further enhance the resolution and context of single-cell analysis, enabling researchers to study chromatin accessibility and spatial organization of gene expression, respectively~\cite{newman2015robust,staahl2016visualization}.
Single-cell analysis has been applied in various fields, such as uncovering tumor heterogeneity in cancer~\cite{levy2016metabolic}, profiling immune cells in health and disease~\cite{villani2017single}, and mapping developmental trajectories of tissues and organs. However, this field faces significant challenges, such as managing large and complex datasets~\cite{wu2020tools,tejada2023causal}, dealing with sparse data~\cite{bouland2023consequences}, and handling the computational needs of large-scale studies~\cite{wolf2018scanpy}. Recent advances in computational tools and algorithms are addressing these issues, allowing for more comprehensive biological insights~\cite{lopez2018deep,eraslan2019single}.

\subsection*{Single-cell foundation models}
The success of foundation models in natural language processing (NLP), such as BERT~\cite{DBLP:conf/naacl/DevlinCLT19} and BART~\cite{DBLP:conf/acl/LewisLGGMLSZ20}, has inspired their application in the single-cell domain, leading to the development of single-cell foundation models with broad applicability across various analytical tasks.
Early approaches to analyzing single-cell gene expression matrices primarily relied on machine learning methods and autoencoder-based frameworks~\cite{liu2021machine,oller2021algorithmic,ji2021machine}. While effective for specific tasks, these models often lacked the generalizability required for a wide range of analyses~\cite{angerer2017single}. 
Recent advances in single-cell foundation models aim to address this limitation. For example, scBERT~\cite{yang2022scbert} leverages the BERT framework to process millions of normalized scRNA-seq datasets, enabling insights into both individual and combinatorial gene expression patterns. Geneformer~\cite{theodoris2023transfer} employs a self-supervised masked token prediction objective to uncover gene networks, with fine-tuning enabling chromatin and network dynamics predictions. scGPT~\cite{cui2024scgpt} utilizes generative pre-training, achieving strong performance in cell type annotation, gene perturbation prediction, and pseudo-cell generation tasks. LangCell~\cite{zhao2024langcell} integrates single-cell data and natural language for unified representation. scFoundation~\cite{hao2024large} leverages self-supervised pre-training on transcriptomics data to learn shared representations of diverse gene expression patterns.

\subsection*{Instruction-following models}

The primary strength of large language models (LLMs) lies in their ability to understand and execute human instructions.  
By training on specialized instruction datasets, these models develop a nuanced understanding of complex tasks, providing greater flexibility and adaptability compared to traditional foundation models.
This capability has driven innovations across biology, including language-guided molecular design~\cite{edwards-etal-2022-translation,fang2023mol,DBLP:journals/corr/abs-2306-11976,tang2024mollm}, medical question-answering~\cite{singhal2023large}, and automated experimental design~\cite{boiko2023autonomous}.
Instruction-following models are now being explored in the single-cell domain. For instance, GPTCelltype~\cite{hou2023reference} demonstrates the feasibility of using GPT-4 for cell type annotation, highlighting the potential of language-guided single-cell analysis. Similarly, Cell2Sentence~\cite{levine2023cell2sentence} translates gene expression profiles into sequences of gene names, showcasing the integration of LLMs into single-cell data analysis.
Apart from them, {\ours} introduces a multi-modal architecture that integrates scRNA-seq data with natural language instructions. Through instruction tuning, it extends LLM capabilities across diverse single-cell analysis tasks.

\section*{Details of multi-modal instruction data}\label{Dataset}

\begin{figure*}[!th] 
  \centering
  \includegraphics[width=0.92\linewidth]{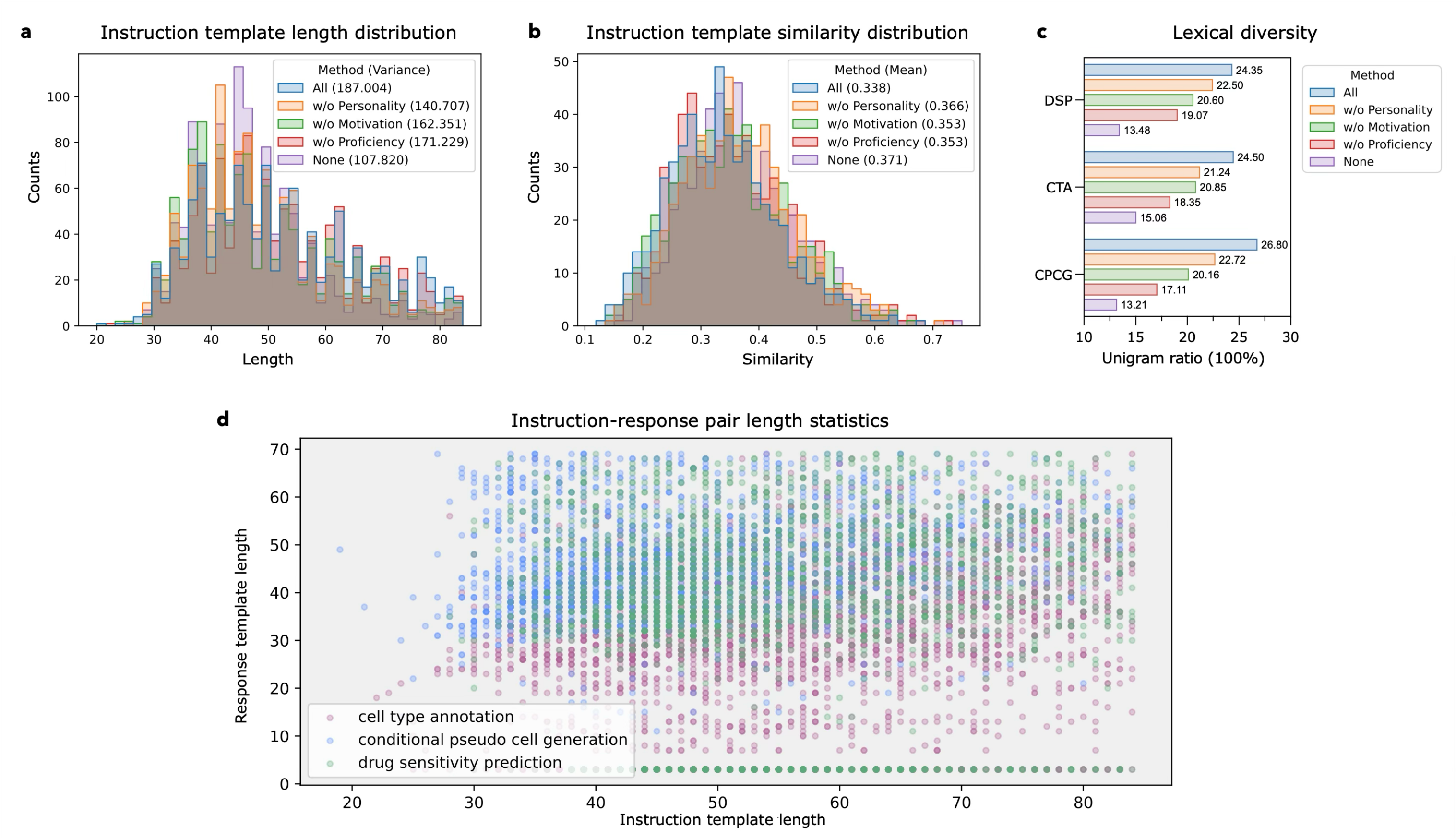}
  \caption{
   \textbf{Statistics of synthetic instruction-response templates.}
   \textbf{a,} Length of instruction templates for different communication styles. Traits such as personality (orange), motivation (green), proficiency (red), and their combination (purple) are systematically removed to evaluate their impact, compared to the full trait version (blue). Variance for each distribution is shown in parentheses.
   \textbf{b,} Similarity of instruction templates across different communication styles. Template similarity is measured pairwise, with samples exceeding a similarity threshold of 0.75 excluded. Average similarity values for each style are reported in parentheses.
   \textbf{c,} Lexical diversity of instruction templates across communication styles. Lexical diversity is quantified using the unigram ratio, defined as the proportion of unique unigrams to the total number of unigrams in each instruction template.
   \textbf{d,} Length distribution of instruction and response templates across tasks. A scatter plot illustrates the lengths of instructions and responses for distinct tasks, with different colors representing task categories.
  }
  \label{fig:statistics}
\end{figure*}

The multi-modal instruction dataset used in our study is pivotal in unlocking the full capabilities of {\ours}. It offers a structured framework that enables the model to effectively navigate the intricate field of single-cell analysis.

Fig.~\ref{fig:statistics} presents a detailed examination of the synthetic templates.
The observed variations in instruction template length and unigram ratio indicate that personality traits significantly influence linguistic style and complexity, resulting in more diverse outputs. Motivation traits also contribute to this diversity, though to a lesser extent. Proficiency traits have a minimal effect on linguistic variation. Notably, a higher mean similarity among samples without personality traits suggests a more standardized, less nuanced language use, underscoring the value of incorporating a range of traits to enhance linguistic richness and adaptability. Such diversity in communication styles is crucial for producing dynamic, effective task-oriented dialogues.

\begin{figure*}[!t] 
  \centering
  \includegraphics[width=0.9\linewidth]{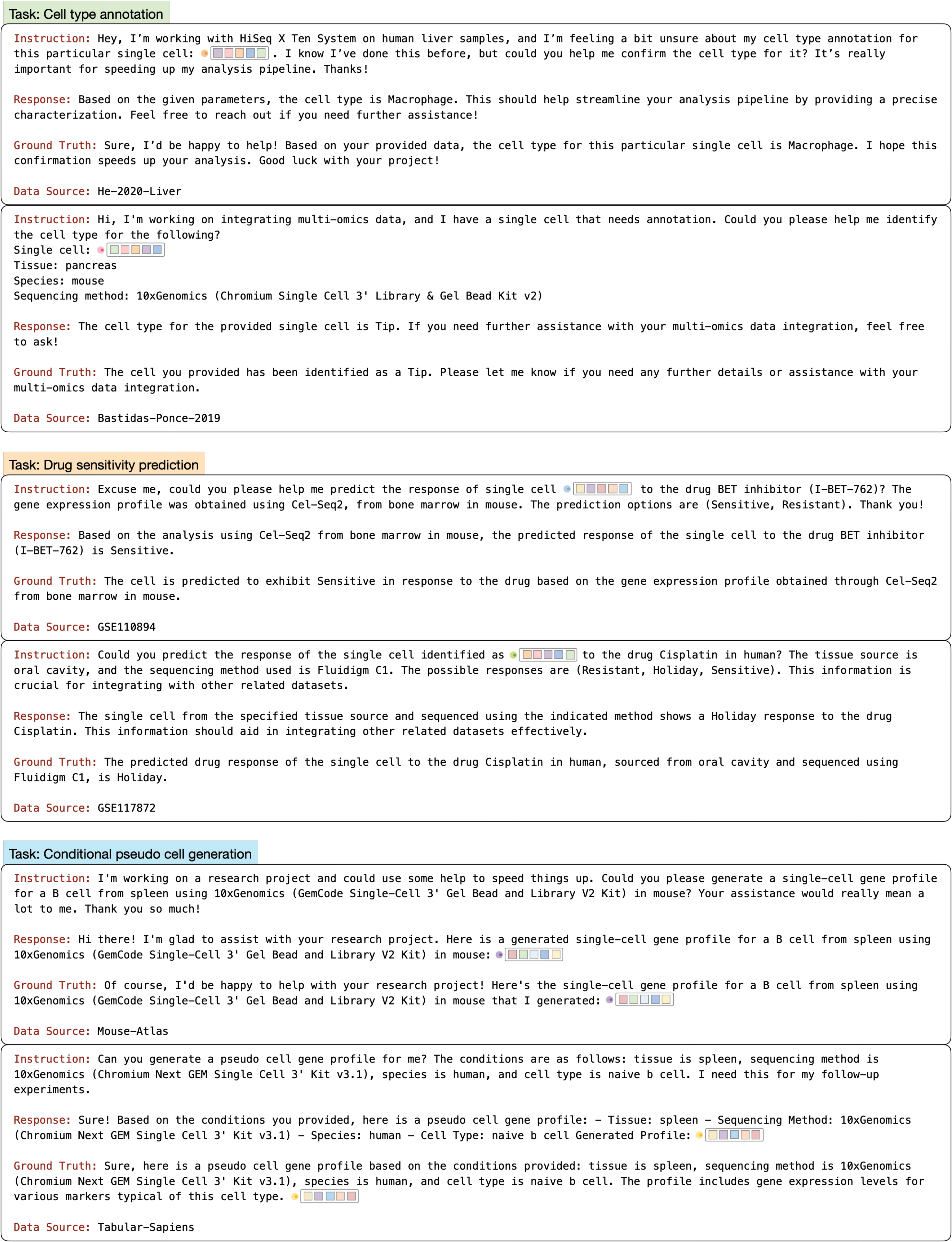}
  \caption{
  \small
  \textbf{Qualitative examples of {\ours}-\textit{chat}.}
  Illustrative examples for each task are presented, showcasing instructions, model-generated responses from {\ours}-\textit{chat}, the corresponding ground truth answers, and their sample sources.
  }
  \label{fig:chat_examples}
\end{figure*}

Fig.~\ref{fig:chat_examples} provides qualitative examples illustrating how {\ours} addresses various single-cell analysis tasks. These examples include the instructions, the model’s responses, corresponding ground-truth answers, and references to the data sources. By directly interfacing with complex biological information, {\ours} simplifies user interactions, enabling life science researchers to efficiently obtain reliable and accurate insights.

\section*{Baselines}\label{Baselines}

To demonstrate the effectiveness of {\ours}, we evaluate its performance in comparison with current state-of-the-art single-cell analysis methods.

\begin{itemize}
    \item \textbf{scGAN}~\cite{marouf2020realistic} is a conditional generative adversarial network tailored for generating realistic single-cell RNA-seq data. It learns complex gene–gene interactions to produce specified cell types, improving downstream tasks like marker gene detection and enhancing algorithm assessments. This approach may also reduce reliance on animal experiments and cut costs.
    \item \textbf{scDiffusion}~\cite{luo2024scdiffusion} is a generative model that combines diffusion and foundation modeling techniques to generate high-quality, condition-controlled scRNA-seq data. It utilizes multiple classifiers to steer the diffusion process for varied condition combinations and introduces Gradient Interpolation—a novel control strategy enabling the generation of continuous cell development trajectories from specified states.
    \item \textbf{Cell2Sentence}~\cite{levine2023cell2sentence} transforms gene expression matrices into expressions using ranked gene names, proving the effectiveness of these cell sentences and establishing a framework for integrating these sentences with GPT-2. This framework involves four tasks: random cell generation, conditional cell generation, cell label prediction, and deriving natural language insights from single-cell data.
    \item \textbf{scBERT}~\cite{yang2022scbert} is a deep neural network-based model that utilizes the bidirectional encoder representations from transformers (BERT) architecture tailored for single-cell analysis. Pretrained on extensive amounts of unlabeled single-cell RNA sequencing (scRNA-seq) data, scBERT captures intricate gene-gene interactions. The model is specifically designed to adapt through fine-tuning to cell type annotation tasks, leveraging its pretrained knowledge to efficiently handle both unseen and user-specific scRNA-seq data in a supervised learning context.
    \item \textbf{scGPT}~\cite{cui2024scgpt} is a single-cell foundation model pre-trained on over 33 million cells and adapts the transformer architecture to learn cell and gene representations simultaneously. This approach establishes a generative pretraining workflow tailored for non-sequential omics data. The model also includes fine-tuning pipelines with task-specific objectives to support a variety of single-cell analysis tasks.
    \item \textbf{Geneformer}~\cite{theodoris2023transfer} is an attention-based deep learning model pretrained on around 30 million single-cell transcriptomes to enable precise predictions in network biology. It captures network dynamics and hierarchy through self-supervised learning during pretraining. Geneformer improves predictive accuracy on various downstream tasks by fine-tuning with limited task-specific data.
\end{itemize}

\subsection*{Baseline experimental setup}

For the comparative analysis in our study, we benchmarked {\ours} against several well-established baseline methods, including scGAN~\cite{marouf2020realistic}, scDiffusion~\cite{luo2024scdiffusion}, Cell2Sentence~\cite{levine2023cell2sentence}, scBERT~\cite{yang2022scbert}, scGPT~\cite{cui2024scgpt}, and Geneformer~\cite{theodoris2023transfer}. 
We reproduced the baseline results by using the code provided in their respective GitHub repositories.
To ensure a fair and consistent evaluation, all models were tested using the same data splits for training, validation, and testing, mirroring our dataset’s configuration. 

\section*{Addtional experimental results}

In this section, we provide further experimental results to highlight the capabilities of {\ours} in single-cell analysis.

\begin{figure*}[!ht] 
  \centering
  \vspace{-0.4cm}
  \includegraphics[width=0.8\linewidth]{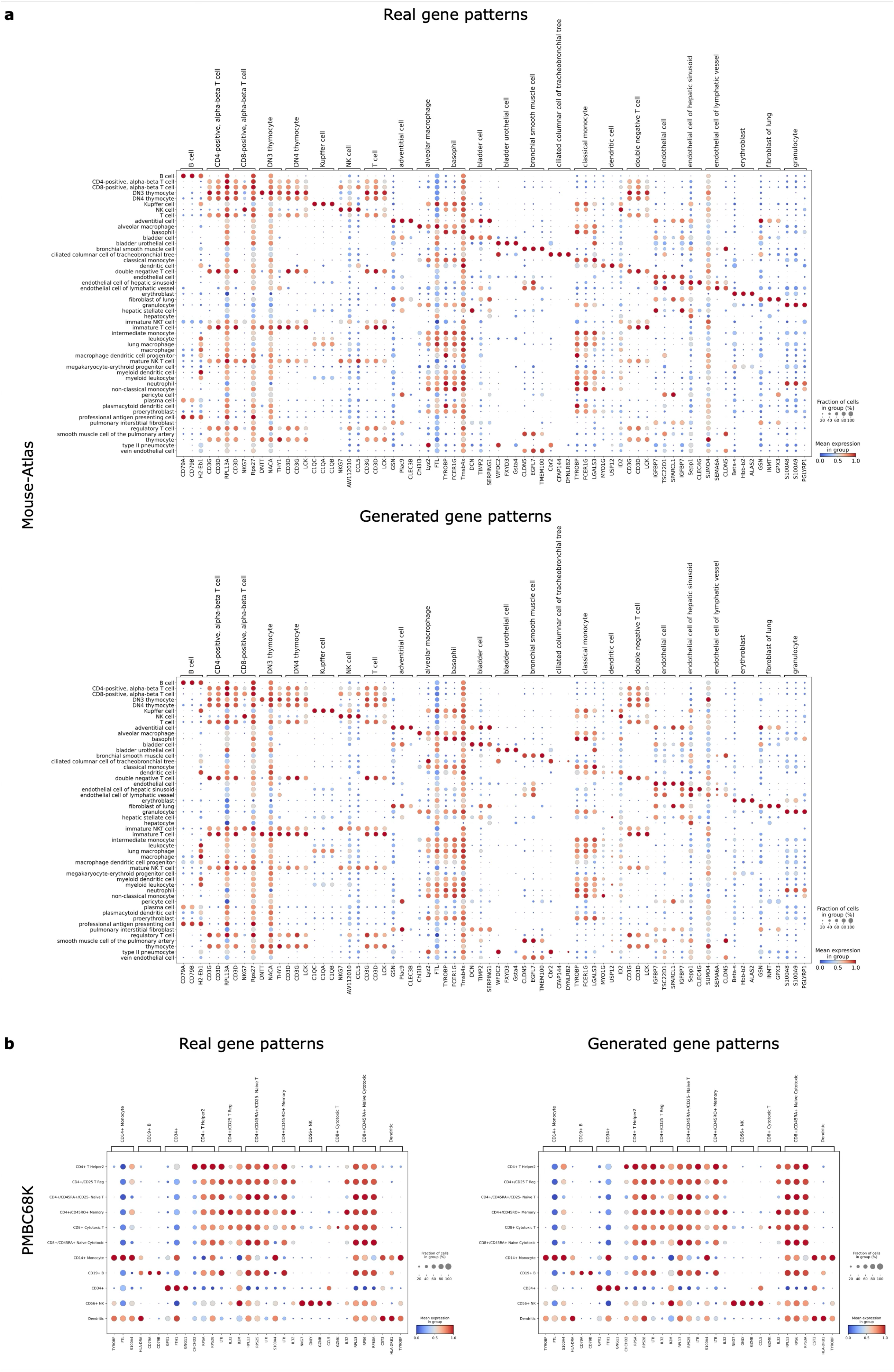}
  \caption{
  \small
  \textbf{Bubble plot for conditional pseudo-cell generation.}
  As a supplement to Fig. 2(b), this plot highlights the top three significant genes for each cell type from the remaining two datasets in our study. 
  For model-generated cells, we display the expression ratios and average expression levels of these significant genes. Redder hues indicate higher average gene expression, while larger circles represent a higher proportion of gene expression within the corresponding cell type.
  }
  \label{fig:bubble_plot}
\end{figure*}

Fig.~\ref{fig:bubble_plot} offers an expanded view on our conditional pseudo-cell generation task by presenting the top three significant genes for each cell type across additional test datasets. The bubble plot visualizes the expression ratios and average expression levels of these genes in the generated cells, offering insights into how closely the model replicates the expected biological profiles.

\begin{figure*}[!ht] 
  \centering
  \vspace{-0.5cm}
  \includegraphics[width=0.8\linewidth]{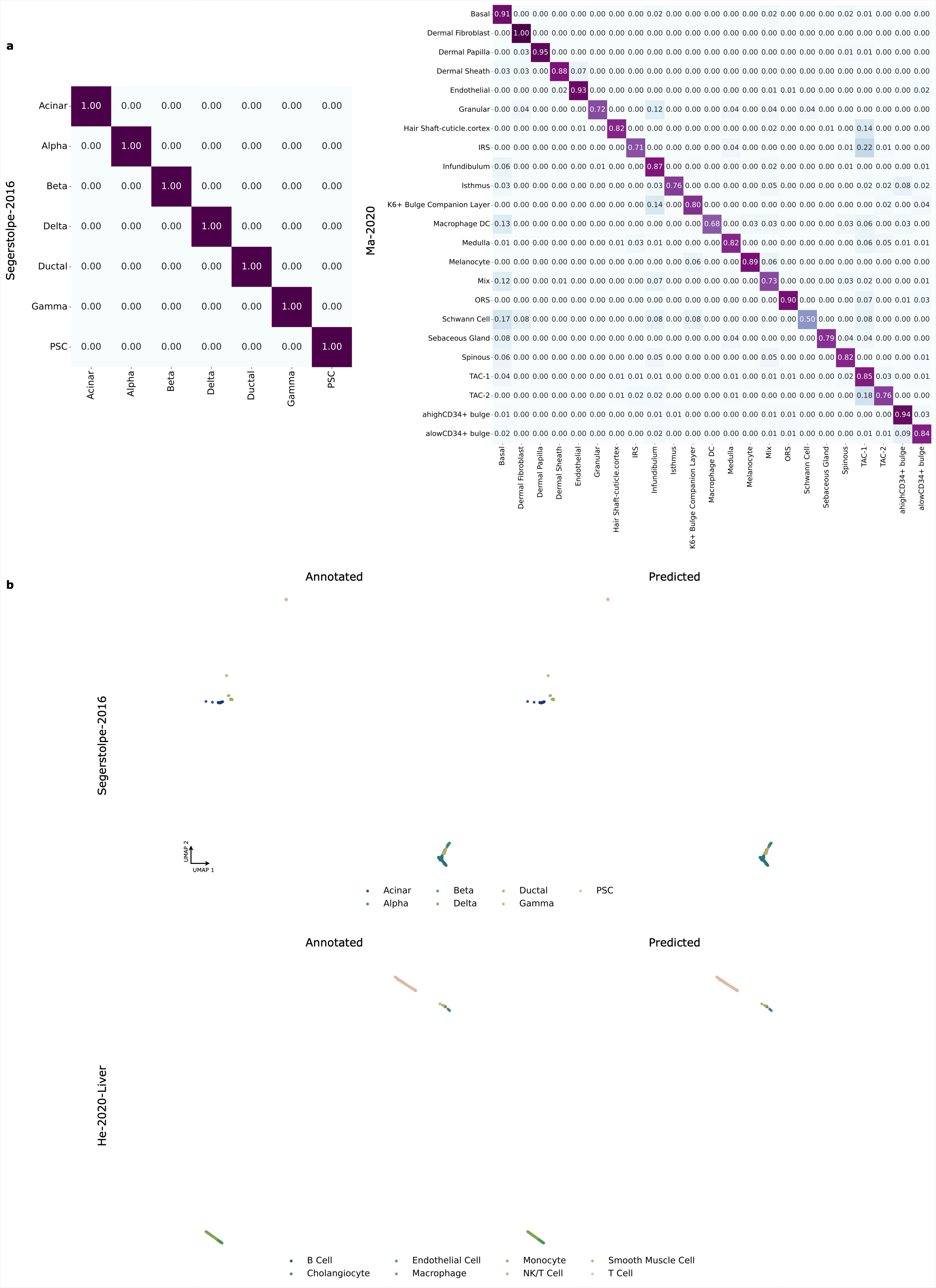}
  \caption{
  \textbf{Confusion matrices and UMAP visualizations for cell type annotation datasets.}
  \textbf{a,} Supplementary confusion matrices for the remaining two datasets, complementing Fig.~3(c). Darker shades represent a higher frequency of agreement between model predictions and the ground-truth cell type annotations. 
  \textbf{b,} UMAP visualizations comparing expert-annotated cell types (left panel) with the predictions generated by {\ours} (right panel). Different colors represent distinct cell types, illustrating the model’s ability to accurately reproduce the structure of the original data.
  }
  \label{fig:confusion_matrix}
\end{figure*}

Fig.~\ref{fig:confusion_matrix} displays confusion matrices and UMAP visualizations for cell type annotation across additional datasets. The confusion matrices demonstrate the agreement between the model’s predictions and ground-truth annotations, with darker shades indicating higher classification accuracy. The accompanying UMAP visualizations provide a qualitative comparison of expert-annotated cell types and the predictions generated by {\ours}. The visual alignment between the two highlights the model’s accuracy in classifying cells based on gene expression data.

\clearpage

\section*{Data availability}

All datasets used in this work are publicly available and were downloaded from the following sources: 
\begin{itemize}
    \item Xin-2016 (\href{https://www.ncbi.nlm.nih.gov/geo/query/acc.cgi?acc=GSE114297}{\textcolor[RGB]{42, 97, 187}{\uline{https://www.ncbi.nlm.nih.gov/geo/query/acc.cgi?acc=GSE114297}}}), 
    \item Segerstolpe-2016 (\href{https://www.ebi.ac.uk/biostudies/arrayexpress/studies/E-MTAB-5061}{\textcolor[RGB]{42, 97, 187}{\uline{https://www.ebi.ac.uk/biostudies/arrayexpress/studies/E-MTAB-5061}}}), 
    \item He-2020 (\href{https://www.ncbi.nlm.nih.gov/geo/query/acc.cgi?acc=GSE159929}{\textcolor[RGB]{42, 97, 187}{\uline{https://www.ncbi.nlm.nih.gov/geo/query/acc.cgi?acc=GSE159929}}}), 
    \item PBMC68K (\href{https://figshare.com/s/49b29cb24b27ec8b6d72}{\textcolor[RGB]{42, 97, 187}{\uline{https://figshare.com/s/49b29cb24b27ec8b6d72}}}), 
    \item GSE117872 (\href{https://www.ncbi.nlm.nih.gov/geo/query/acc.cgi?acc=GSE117872}{\textcolor[RGB]{42, 97, 187}{\uline{https://www.ncbi.nlm.nih.gov/geo/query/acc.cgi?acc=GSE117872}}}), 
    \item GSE149383 (\href{https://github.com/OSU-BMBL/scDEAL}{\textcolor[RGB]{42, 97, 187}{\uline{https://github.com/OSU-BMBL/scDEAL}}}), 
    \item Ma-2020 (\href{https://www.ncbi.nlm.nih.gov/geo/query/acc.cgi?acc=GSE140203}{\textcolor[RGB]{42, 97, 187}{\uline{https://www.ncbi.nlm.nih.gov/geo/query/acc.cgi?acc=GSE140203}}}), 
    \item Bastidas-Ponce-2019 (\href{https://www.ncbi.nlm.nih.gov/geo/query/acc.cgi?acc=GSE132188}{\textcolor[RGB]{42, 97, 187}{\uline{https://www.ncbi.nlm.nih.gov/geo/query/acc.cgi?acc=GSE132188}}}), 
    \item GSE110894 (\href{https://www.ncbi.nlm.nih.gov/geo/query/acc.cgi?acc=GSE110894}{\textcolor[RGB]{42, 97, 187}{\uline{https://www.ncbi.nlm.nih.gov/geo/query/acc.cgi?acc=GSE110894}}}), 
    \item Mouse-Atlas (\href{https://www.ncbi.nlm.nih.gov/geo/query/acc.cgi?acc=GSM4505404}{\textcolor[RGB]{42, 97, 187}{\uline{https://www.ncbi.nlm.nih.gov/geo/query/acc.cgi?acc=GSM4505404}}}).
\end{itemize}

\section*{Code availability}
The source code of this work is freely available on GitHub at \href{https://github.com/zjunlp/InstructCell}{\textcolor[RGB]{42, 97, 187}{\uline{https://github.com/zjunlp/InstructCell}}}. 

The models developed in this work are accessible on Hugging Face at:
\begin{itemize}
    \item zjunlp/InstructCell-chat (\href{https://huggingface.co/zjunlp/InstructCell-chat}{\textcolor[RGB]{42, 97, 187}{\uline{https://huggingface.co/zjunlp/InstructCell-chat}}})
    \item zjunlp/InstructCell-instruct (\href{https://huggingface.co/zjunlp/InstructCell-instruct}{\textcolor[RGB]{42, 97, 187}{\uline{https://huggingface.co/zjunlp/InstructCell-instruct}}})
\end{itemize}

\newpage

\normalem
\bibliographystyle{unsrt}
\bibliography{ic}

\begin{thebibliography}{100}

\bibitem{gpt4}
GPT-4.
\newblock \url{https://openai.com/gpt-4}, Accessed 15 Apr 2024.

\bibitem{DBLP:journals/jmlr/ChowdheryNDBMRBCSGSSTMRBTSPRDHPBAI23}
Aakanksha Chowdhery, Sharan Narang, Jacob Devlin, Maarten Bosma, Gaurav Mishra, Adam Roberts, Paul Barham, Hyung~Won Chung, Charles Sutton, Sebastian Gehrmann, Parker Schuh, Kensen Shi, Sasha Tsvyashchenko, Joshua Maynez, Abhishek Rao, Parker Barnes, Yi~Tay, Noam Shazeer, Vinodkumar Prabhakaran, Emily Reif, Nan Du, Ben Hutchinson, Reiner Pope, James Bradbury, Jacob Austin, Michael Isard, Guy Gur{-}Ari, Pengcheng Yin, Toju Duke, Anselm Levskaya, Sanjay Ghemawat, Sunipa Dev, Henryk Michalewski, Xavier Garcia, Vedant Misra, Kevin Robinson, Liam Fedus, Denny Zhou, Daphne Ippolito, David Luan, Hyeontaek Lim, Barret Zoph, Alexander Spiridonov, Ryan Sepassi, David Dohan, Shivani Agrawal, Mark Omernick, Andrew~M. Dai, Thanumalayan~Sankaranarayana Pillai, Marie Pellat, Aitor Lewkowycz, Erica Moreira, Rewon Child, Oleksandr Polozov, Katherine Lee, Zongwei Zhou, Xuezhi Wang, Brennan Saeta, Mark Diaz, Orhan Firat, Michele Catasta, Jason Wei, Kathy Meier{-}Hellstern, Douglas Eck, Jeff Dean, Slav Petrov, and Noah Fiedel.
\newblock Palm: Scaling language modeling with pathways.
\newblock {\em J. Mach. Learn. Res.}, 24:240:1--240:113, 2023.

\bibitem{DBLP:journals/corr/abs-2302-13971}
Hugo Touvron, Thibaut Lavril, Gautier Izacard, Xavier Martinet, Marie{-}Anne Lachaux, Timoth{\'{e}}e Lacroix, Baptiste Rozi{\`{e}}re, Naman Goyal, Eric Hambro, Faisal Azhar, Aur{\'{e}}lien Rodriguez, Armand Joulin, Edouard Grave, and Guillaume Lample.
\newblock Llama: Open and efficient foundation language models.
\newblock {\em CoRR}, abs/2302.13971, 2023.

\bibitem{claude}
Claude anthropic.
\newblock \url{https://www.anthropic.com/claude}, Accessed 27 Jun 2024.

\bibitem{DBLP:conf/nips/Ouyang0JAWMZASR22}
Long Ouyang, Jeffrey Wu, Xu~Jiang, Diogo Almeida, Carroll~L. Wainwright, Pamela Mishkin, Chong Zhang, Sandhini Agarwal, Katarina Slama, Alex Ray, John Schulman, Jacob Hilton, Fraser Kelton, Luke Miller, Maddie Simens, Amanda Askell, Peter Welinder, Paul~F. Christiano, Jan Leike, and Ryan Lowe.
\newblock Training language models to follow instructions with human feedback.
\newblock In {\em NeurIPS}, 2022.

\bibitem{DBLP:conf/iclr/SanhWRBSACSRDBX22}
Victor Sanh, Albert Webson, Colin Raffel, Stephen~H. Bach, Lintang Sutawika, Zaid Alyafeai, Antoine Chaffin, Arnaud Stiegler, Arun Raja, Manan Dey, M~Saiful Bari, Canwen Xu, Urmish Thakker, Shanya~Sharma Sharma, Eliza Szczechla, Taewoon Kim, Gunjan Chhablani, Nihal~V. Nayak, Debajyoti Datta, Jonathan Chang, Mike~Tian{-}Jian Jiang, Han Wang, Matteo Manica, Sheng Shen, Zheng~Xin Yong, Harshit Pandey, Rachel Bawden, Thomas Wang, Trishala Neeraj, Jos Rozen, Abheesht Sharma, Andrea Santilli, Thibault F{\'{e}}vry, Jason~Alan Fries, Ryan Teehan, Teven~Le Scao, Stella Biderman, Leo Gao, Thomas Wolf, and Alexander~M. Rush.
\newblock Multitask prompted training enables zero-shot task generalization.
\newblock In {\em {ICLR}}. OpenReview.net, 2022.

\bibitem{brazma2000gene}
Alvis Brazma and Jaak Vilo.
\newblock Gene expression data analysis.
\newblock {\em FEBS letters}, 480(1):17--24, 2000.

\bibitem{plass2018cell}
Mireya Plass, Jordi Solana, F~Alexander Wolf, Salah Ayoub, Aristotelis Misios, Petar Gla{\v{z}}ar, Benedikt Obermayer, Fabian~J Theis, Christine Kocks, and Nikolaus Rajewsky.
\newblock Cell type atlas and lineage tree of a whole complex animal by single-cell transcriptomics.
\newblock {\em Science}, 360(6391):eaaq1723, 2018.

\bibitem{cao2019single}
Junyue Cao, Malte Spielmann, Xiaojie Qiu, Xingfan Huang, Daniel~M Ibrahim, Andrew~J Hill, Fan Zhang, Stefan Mundlos, Lena Christiansen, Frank~J Steemers, et~al.
\newblock The single-cell transcriptional landscape of mammalian organogenesis.
\newblock {\em Nature}, 566(7745):496--502, 2019.

\bibitem{barrett2012ncbi}
Tanya Barrett, Stephen~E Wilhite, Pierre Ledoux, Carlos Evangelista, Irene~F Kim, Maxim Tomashevsky, Kimberly~A Marshall, Katherine~H Phillippy, Patti~M Sherman, Michelle Holko, et~al.
\newblock Ncbi geo: archive for functional genomics data sets—update.
\newblock {\em Nucleic acids research}, 41(D1):D991--D995, 2012.

\bibitem{regev2017human}
Aviv Regev, Sarah~A Teichmann, Eric~S Lander, Ido Amit, Christophe Benoist, Ewan Birney, Bernd Bodenmiller, Peter Campbell, Piero Carninci, Menna Clatworthy, et~al.
\newblock The human cell atlas.
\newblock {\em elife}, 6:e27041, 2017.

\bibitem{yang2022scbert}
Fan Yang, Wenchuan Wang, Fang Wang, Yuan Fang, Duyu Tang, Junzhou Huang, Hui Lu, and Jianhua Yao.
\newblock scbert as a large-scale pretrained deep language model for cell type annotation of single-cell rna-seq data.
\newblock {\em Nature Machine Intelligence}, 4(10):852--866, 2022.

\bibitem{theodoris2023transfer}
Christina~V Theodoris, Ling Xiao, Anant Chopra, Mark~D Chaffin, Zeina~R Al~Sayed, Matthew~C Hill, Helene Mantineo, Elizabeth~M Brydon, Zexian Zeng, X~Shirley Liu, et~al.
\newblock Transfer learning enables predictions in network biology.
\newblock {\em Nature}, 618(7965):616--624, 2023.

\bibitem{cui2024scgpt}
Haotian Cui, Chloe Wang, Hassaan Maan, Kuan Pang, Fengning Luo, Nan Duan, and Bo~Wang.
\newblock scgpt: toward building a foundation model for single-cell multi-omics using generative ai.
\newblock {\em Nature Methods}, pages 1--11, 2024.

\bibitem{hao2024large}
Minsheng Hao, Jing Gong, Xin Zeng, Chiming Liu, Yucheng Guo, Xingyi Cheng, Taifeng Wang, Jianzhu Ma, Xuegong Zhang, and Le~Song.
\newblock Large-scale foundation model on single-cell transcriptomics.
\newblock {\em Nature Methods}, pages 1--11, 2024.

\bibitem{levine2023cell2sentence}
Daniel LeVine, Syed~Asad Rizvi, Sacha L{\'{e}}vy, Nazreen Pallikkavaliyaveetil, David Zhang, Xingyu Chen, Sina Ghadermarzi, Ruiming Wu, Zihe Zheng, Ivan Vrkic, Anna Zhong, Daphne Raskin, Insu Han, Antonio~Henrique de~Oliveira~Fonseca, Josue~Ortega Caro, Amin Karbasi, Rahul~Madhav Dhodapkar, and David van Dijk.
\newblock Cell2sentence: Teaching large language models the language of biology.
\newblock In {\em {ICML}}. OpenReview.net, 2024.

\bibitem{hou2023reference}
Wenpin Hou and Zhicheng Ji.
\newblock Assessing gpt-4 for cell type annotation in single-cell rna-seq analysis.
\newblock {\em Nature Methods}, pages 1--4, 2024.

\bibitem{chen2023genept}
Yiqun Chen and James Zou.
\newblock Simple and effective embedding model for single-cell biology built from chatgpt.
\newblock {\em Nature Biomedical Engineering}, pages 1--11, 2024.

\bibitem{liu2023scelmo}
Tianyu Liu, Tianqi Chen, Wangjie Zheng, Xiao Luo, and Hongyu Zhao.
\newblock scelmo: Embeddings from language models are good learners for single-cell data analysis.
\newblock {\em bioRxiv}, pages 2023--12, 2023.

\bibitem{gpt4o}
GPT-4o.
\newblock \url{https://openai.com/index/hello-gpt-4o}, Accessed 29 May 2024.

\bibitem{DBLP:conf/emnlp/DingCXQHL0Z23}
Ning Ding, Yulin Chen, Bokai Xu, Yujia Qin, Shengding Hu, Zhiyuan Liu, Maosong Sun, and Bowen Zhou.
\newblock Enhancing chat language models by scaling high-quality instructional conversations.
\newblock In {\em {EMNLP}}, pages 3029--3051. Association for Computational Linguistics, 2023.

\bibitem{DBLP:conf/icml/0008LSH23}
Junnan Li, Dongxu Li, Silvio Savarese, and Steven C.~H. Hoi.
\newblock {BLIP-2:} bootstrapping language-image pre-training with frozen image encoders and large language models.
\newblock In {\em {ICML}}, volume 202 of {\em Proceedings of Machine Learning Research}, pages 19730--19742. {PMLR}, 2023.

\bibitem{radford2019language}
Alec Radford, Jeffrey Wu, Rewon Child, David Luan, Dario Amodei, Ilya Sutskever, et~al.
\newblock Language models are unsupervised multitask learners.
\newblock {\em OpenAI blog}, 1(8):9, 2019.

\bibitem{DBLP:conf/acl/LewisLGGMLSZ20}
Mike Lewis, Yinhan Liu, Naman Goyal, Marjan Ghazvininejad, Abdelrahman Mohamed, Omer Levy, Veselin Stoyanov, and Luke Zettlemoyer.
\newblock {BART:} denoising sequence-to-sequence pre-training for natural language generation, translation, and comprehension.
\newblock In {\em {ACL}}, pages 7871--7880. Association for Computational Linguistics, 2020.

\bibitem{DBLP:journals/jmlr/RaffelSRLNMZLL20}
Colin Raffel, Noam Shazeer, Adam Roberts, Katherine Lee, Sharan Narang, Michael Matena, Yanqi Zhou, Wei Li, and Peter~J. Liu.
\newblock Exploring the limits of transfer learning with a unified text-to-text transformer.
\newblock {\em J. Mach. Learn. Res.}, 21:140:1--140:67, 2020.

\bibitem{wu2024towards}
Haoning Wu, Hanwei Zhu, Zicheng Zhang, Erli Zhang, Chaofeng Chen, Liang Liao, Chunyi Li, Annan Wang, Wenxiu Sun, Qiong Yan, Xiaohong Liu, Guangtao Zhai, Shiqi Wang, and Weisi Lin.
\newblock Towards open-ended visual quality comparison.
\newblock In {\em {ECCV} {(3)}}, volume 15061 of {\em Lecture Notes in Computer Science}, pages 360--377. Springer, 2024.

\bibitem{jiang2024mantis}
Dongfu Jiang, Xuan He, Huaye Zeng, Cong Wei, Max Ku, Qian Liu, and Wenhu Chen.
\newblock Mantis: Interleaved multi-image instruction tuning.
\newblock {\em arXiv preprint arXiv:2405.01483}, 2024.

\bibitem{wu2023next}
Shengqiong Wu, Hao Fei, Leigang Qu, Wei Ji, and Tat-Seng Chua.
\newblock Next-gpt: Any-to-any multimodal llm.
\newblock {\em arXiv preprint arXiv:2309.05519}, 2023.

\bibitem{DBLP:conf/nips/KohFS23}
Jing~Yu Koh, Daniel Fried, and Russ Salakhutdinov.
\newblock Generating images with multimodal language models.
\newblock In {\em NeurIPS}, 2023.

\bibitem{DBLP:conf/nips/SohnLY15}
Kihyuk Sohn, Honglak Lee, and Xinchen Yan.
\newblock Learning structured output representation using deep conditional generative models.
\newblock In {\em {NIPS}}, pages 3483--3491, 2015.

\bibitem{welch1947generalization}
Bernard~L Welch.
\newblock The generalization of ‘student's’problem when several different population varlances are involved.
\newblock {\em Biometrika}, 34(1-2):28--35, 1947.

\bibitem{luo2024scdiffusion}
Erpai Luo, Minsheng Hao, Lei Wei, and Xuegong Zhang.
\newblock scdiffusion: conditional generation of high-quality single-cell data using diffusion model.
\newblock {\em Bioinformatics}, 40(9):btae518, 2024.

\bibitem{marouf2020realistic}
Mohamed Marouf, Pierre Machart, Vikas Bansal, Christoph Kilian, Daniel~S Magruder, Christian~F Krebs, and Stefan Bonn.
\newblock Realistic in silico generation and augmentation of single-cell rna-seq data using generative adversarial networks.
\newblock {\em Nature communications}, 11(1):166, 2020.

\bibitem{DBLP:journals/corr/SimonyanVZ13}
Karen Simonyan, Andrea Vedaldi, and Andrew Zisserman.
\newblock Deep inside convolutional networks: Visualising image classification models and saliency maps.
\newblock In {\em {ICLR} (Workshop Poster)}, 2014.

\bibitem{DBLP:journals/nar/HuLXZLBCJYOLWZ23}
Congxue Hu, Tengyue Li, Yingqi Xu, Xinxin Zhang, Feng Li, Jing Bai, Jing Chen, Wenqi Jiang, Kaiyue Yang, Qi~Ou, Xia Li, Peng Wang, and Yunpeng Zhang.
\newblock Cellmarker 2.0: an updated database of manually curated cell markers in human/mouse and web tools based on scrna-seq data.
\newblock {\em Nucleic Acids Res.}, 51({D1}):870--876, 2023.

\bibitem{muraro2016single}
Mauro~J Muraro, Gitanjali Dharmadhikari, Dominic Gr{\"u}n, Nathalie Groen, Tim Dielen, Erik Jansen, Leon Van~Gurp, Marten~A Engelse, Francoise Carlotti, Eelco~Jp De~Koning, et~al.
\newblock A single-cell transcriptome atlas of the human pancreas.
\newblock {\em Cell systems}, 3(4):385--394, 2016.

\bibitem{van2022generation}
L{\'e}on van Gurp, Leon Fodoulian, Daniel Oropeza, Kenichiro Furuyama, Eva Bru-Tari, Anh~Nguyet Vu, John~S Kaddis, Iv{\'a}n Rodr{\'\i}guez, Fabrizio Thorel, and Pedro~L Herrera.
\newblock Generation of human islet cell type-specific identity genesets.
\newblock {\em Nature communications}, 13(1):2020, 2022.

\bibitem{DBLP:conf/nips/ZhengC00WZL0LXZ23}
Lianmin Zheng, Wei{-}Lin Chiang, Ying Sheng, Siyuan Zhuang, Zhanghao Wu, Yonghao Zhuang, Zi~Lin, Zhuohan Li, Dacheng Li, Eric~P. Xing, Hao Zhang, Joseph~E. Gonzalez, and Ion Stoica.
\newblock Judging llm-as-a-judge with mt-bench and chatbot arena.
\newblock In {\em NeurIPS}, 2023.

\bibitem{DBLP:journals/corr/abs-2311-09766}
Yiqi Liu, Nafise~Sadat Moosavi, and Chenghua Lin.
\newblock Llms as narcissistic evaluators: When ego inflates evaluation scores.
\newblock {\em CoRR}, abs/2311.09766, 2023.

\bibitem{DBLP:conf/pakdd/AwalCLM21}
Md.~Rabiul Awal, Rui Cao, Roy~Ka{-}Wei Lee, and Sandra Mitrovic.
\newblock Angrybert: Joint learning target and emotion for hate speech detection.
\newblock In {\em {PAKDD} {(1)}}, volume 12712 of {\em Lecture Notes in Computer Science}, pages 701--713. Springer, 2021.

\bibitem{DBLP:journals/tkde/ZhangY22}
Yu~Zhang and Qiang Yang.
\newblock A survey on multi-task learning.
\newblock {\em {IEEE} Trans. Knowl. Data Eng.}, 34(12):5586--5609, 2022.

\bibitem{roohani2024predicting}
Yusuf Roohani, Kexin Huang, and Jure Leskovec.
\newblock Predicting transcriptional outcomes of novel multigene perturbations with gears.
\newblock {\em Nature Biotechnology}, 42(6):927--935, 2024.

\bibitem{wei2021finetuned}
Jason Wei, Maarten Bosma, Vincent~Y Zhao, Kelvin Guu, Adams~Wei Yu, Brian Lester, Nan Du, Andrew~M Dai, and Quoc~V Le.
\newblock Finetuned language models are zero-shot learners.
\newblock {\em arXiv preprint arXiv:2109.01652}, 2021.

\bibitem{chung2024scaling}
Hyung~Won Chung, Le~Hou, Shayne Longpre, Barret Zoph, Yi~Tay, William Fedus, Yunxuan Li, Xuezhi Wang, Mostafa Dehghani, Siddhartha Brahma, et~al.
\newblock Scaling instruction-finetuned language models.
\newblock {\em Journal of Machine Learning Research}, 25(70):1--53, 2024.

\bibitem{he2024zero}
Bingxiang He, Ning Ding, Cheng Qian, Jia Deng, Ganqu Cui, Lifan Yuan, Huan-ang Gao, Huimin Chen, Zhiyuan Liu, and Maosong Sun.
\newblock Zero-shot generalization during instruction tuning: Insights from similarity and granularity.
\newblock {\em arXiv preprint arXiv:2406.11721}, 2024.

\bibitem{ashuach2022peakvi}
Tal Ashuach, Daniel~A Reidenbach, Adam Gayoso, and Nir Yosef.
\newblock Peakvi: A deep generative model for single-cell chromatin accessibility analysis.
\newblock {\em Cell reports methods}, 2(3), 2022.

\bibitem{liu2024muse}
Tianyu Liu, Yuge Wang, Rex Ying, and Hongyu Zhao.
\newblock Muse-gnn: learning unified gene representation from multimodal biological graph data.
\newblock {\em Advances in neural information processing systems}, 36, 2024.

\bibitem{lin2004rouge}
Chin-Yew Lin.
\newblock Rouge: A package for automatic evaluation of summaries.
\newblock In {\em Text summarization branches out}, pages 74--81, 2004.

\bibitem{vaswani2017attention}
Ashish Vaswani.
\newblock Attention is all you need.
\newblock {\em arXiv preprint arXiv:1706.03762}, 2017.

\bibitem{he2016deep}
Kaiming He, Xiangyu Zhang, Shaoqing Ren, and Jian Sun.
\newblock Deep residual learning for image recognition.
\newblock In {\em Proceedings of the IEEE conference on computer vision and pattern recognition}, pages 770--778, 2016.

\bibitem{li2023blip}
Junnan Li, Dongxu Li, Silvio Savarese, and Steven Hoi.
\newblock Blip-2: Bootstrapping language-image pre-training with frozen image encoders and large language models.
\newblock In {\em International conference on machine learning}, pages 19730--19742. PMLR, 2023.

\bibitem{kingma2013auto}
DP~Kingma.
\newblock Auto-encoding variational bayes.
\newblock {\em arXiv preprint arXiv:1312.6114}, 2013.

\bibitem{love2014moderated}
Michael~I Love, Wolfgang Huber, and Simon Anders.
\newblock Moderated estimation of fold change and dispersion for rna-seq data with deseq2.
\newblock {\em Genome biology}, 15:1--21, 2014.

\bibitem{choudhary2022comparison}
Saket Choudhary and Rahul Satija.
\newblock Comparison and evaluation of statistical error models for scrna-seq.
\newblock {\em Genome biology}, 23(1):27, 2022.

\bibitem{ding2020systematic}
Jiarui Ding, Xian Adiconis, Sean~K Simmons, Monika~S Kowalczyk, Cynthia~C Hession, Nemanja~D Marjanovic, Travis~K Hughes, Marc~H Wadsworth, Tyler Burks, Lan~T Nguyen, et~al.
\newblock Systematic comparison of single-cell and single-nucleus rna-sequencing methods.
\newblock {\em Nature biotechnology}, 38(6):737--746, 2020.

\bibitem{bouland2023consequences}
Gerard~A Bouland, Ahmed Mahfouz, and Marcel~JT Reinders.
\newblock Consequences and opportunities arising due to sparser single-cell rna-seq datasets.
\newblock {\em Genome biology}, 24(1):86, 2023.

\bibitem{vieth2017powsimr}
Beate Vieth, Christoph Ziegenhain, Swati Parekh, Wolfgang Enard, and Ines Hellmann.
\newblock powsimr: power analysis for bulk and single cell rna-seq experiments.
\newblock {\em Bioinformatics}, 33(21):3486--3488, 2017.

\bibitem{eraslan2019single}
G{\"o}kcen Eraslan, Lukas~M Simon, Maria Mircea, Nikola~S Mueller, and Fabian~J Theis.
\newblock Single-cell rna-seq denoising using a deep count autoencoder.
\newblock {\em Nature communications}, 10(1):390, 2019.

\bibitem{lopez2018deep}
Romain Lopez, Jeffrey Regier, Michael~B Cole, Michael~I Jordan, and Nir Yosef.
\newblock Deep generative modeling for single-cell transcriptomics.
\newblock {\em Nature methods}, 15(12):1053--1058, 2018.

\bibitem{shao2020controlvae}
Huajie Shao, Shuochao Yao, Dachun Sun, Aston Zhang, Shengzhong Liu, Dongxin Liu, Jun Wang, and Tarek Abdelzaher.
\newblock Controlvae: Controllable variational autoencoder.
\newblock In {\em International conference on machine learning}, pages 8655--8664. PMLR, 2020.

\bibitem{ho2022classifier}
Jonathan Ho and Tim Salimans.
\newblock Classifier-free diffusion guidance.
\newblock {\em arXiv preprint arXiv:2207.12598}, 2022.

\bibitem{devlin2018bert}
Jacob Devlin, Ming{-}Wei Chang, Kenton Lee, and Kristina Toutanova.
\newblock {BERT:} pre-training of deep bidirectional transformers for language understanding.
\newblock In {\em {NAACL-HLT} {(1)}}, pages 4171--4186. Association for Computational Linguistics, 2019.

\bibitem{liu2019roberta}
Yinhan Liu, Myle Ott, Naman Goyal, Jingfei Du, Mandar Joshi, Danqi Chen, Omer Levy, Mike Lewis, Luke Zettlemoyer, and Veselin Stoyanov.
\newblock Roberta: A robustly optimized bert pretraining approach.
\newblock {\em arXiv preprint arXiv:1907.11692}, 2019.

\bibitem{lan2019albert}
Zhenzhong Lan, Mingda Chen, Sebastian Goodman, Kevin Gimpel, Piyush Sharma, and Radu Soricut.
\newblock {ALBERT:} {A} lite {BERT} for self-supervised learning of language representations.
\newblock In {\em {ICLR}}. OpenReview.net, 2020.

\bibitem{he2020deberta}
Pengcheng He, Xiaodong Liu, Jianfeng Gao, and Weizhu Chen.
\newblock Deberta: decoding-enhanced bert with disentangled attention.
\newblock In {\em {ICLR}}. OpenReview.net, 2021.

\bibitem{radford2018improving}
Alec Radford, Karthik Narasimhan, Tim Salimans, Ilya Sutskever, et~al.
\newblock Improving language understanding by generative pre-training.
\newblock 2018.

\bibitem{brown2020language}
Tom Brown, Benjamin Mann, Nick Ryder, Melanie Subbiah, Jared~D Kaplan, Prafulla Dhariwal, Arvind Neelakantan, Pranav Shyam, Girish Sastry, Amanda Askell, et~al.
\newblock Language models are few-shot learners.
\newblock {\em Advances in neural information processing systems}, 33:1877--1901, 2020.

\bibitem{wang2021gpt}
Ben Wang and Aran Komatsuzaki.
\newblock Gpt-j-6b: A 6 billion parameter autoregressive language model, 2021.

\bibitem{DBLP:conf/nips/00040WWLWGZH19}
Li~Dong, Nan Yang, Wenhui Wang, Furu Wei, Xiaodong Liu, Yu~Wang, Jianfeng Gao, Ming Zhou, and Hsiao{-}Wuen Hon.
\newblock Unified language model pre-training for natural language understanding and generation.
\newblock In {\em NeurIPS}, pages 13042--13054, 2019.

\bibitem{raffel2020exploring}
Colin Raffel, Noam Shazeer, Adam Roberts, Katherine Lee, Sharan Narang, Michael Matena, Yanqi Zhou, Wei Li, and Peter~J Liu.
\newblock Exploring the limits of transfer learning with a unified text-to-text transformer.
\newblock {\em Journal of machine learning research}, 21(140):1--67, 2020.

\bibitem{zhang2020pegasus}
Jingqing Zhang, Yao Zhao, Mohammad Saleh, and Peter Liu.
\newblock Pegasus: Pre-training with extracted gap-sentences for abstractive summarization.
\newblock In {\em International conference on machine learning}, pages 11328--11339. PMLR, 2020.

\bibitem{baxter1997bayesian}
Jonathan Baxter.
\newblock A bayesian/information theoretic model of learning to learn via multiple task sampling.
\newblock {\em Machine learning}, 28:7--39, 1997.

\bibitem{ruder2017overview}
Sebastian Ruder.
\newblock An overview of multi-task learning in deep neural networks.
\newblock {\em arXiv preprint arXiv:1706.05098}, 2017.

\bibitem{DBLP:journals/corr/abs-2405-11874}
Qingchen Yu, Zifan Zheng, Shichao Song, Zhiyu Li, Feiyu Xiong, Bo~Tang, and Ding Chen.
\newblock xfinder: Robust and pinpoint answer extraction for large language models.
\newblock {\em CoRR}, abs/2405.11874, 2024.

\bibitem{mcinnes2018umap}
Leland McInnes, John Healy, and James Melville.
\newblock Umap: Uniform manifold approximation and projection for dimension reduction.
\newblock {\em arXiv preprint arXiv:1802.03426}, 2018.

\bibitem{gretton2012kernel}
Arthur Gretton, Karsten~M Borgwardt, Malte~J Rasch, Bernhard Sch{\"o}lkopf, and Alexander Smola.
\newblock A kernel two-sample test.
\newblock {\em The Journal of Machine Learning Research}, 13(1):723--773, 2012.

\bibitem{shaham2017removal}
Uri Shaham, Kelly~P Stanton, Jun Zhao, Huamin Li, Khadir Raddassi, Ruth Montgomery, and Yuval Kluger.
\newblock Removal of batch effects using distribution-matching residual networks.
\newblock {\em Bioinformatics}, 33(16):2539--2546, 2017.

\bibitem{DBLP:conf/nips/AdebayoGMGHK18}
Julius Adebayo, Justin Gilmer, Michael Muelly, Ian~J. Goodfellow, Moritz Hardt, and Been Kim.
\newblock Sanity checks for saliency maps.
\newblock In {\em NeurIPS}, pages 9525--9536, 2018.

\bibitem{van2008visualizing}
Laurens Van~der Maaten and Geoffrey Hinton.
\newblock Visualizing data using t-sne.
\newblock {\em Journal of machine learning research}, 9(11), 2008.

\bibitem{becht2019dimensionality}
Etienne Becht, Leland McInnes, John Healy, Charles-Antoine Dutertre, Immanuel~WH Kwok, Lai~Guan Ng, Florent Ginhoux, and Evan~W Newell.
\newblock Dimensionality reduction for visualizing single-cell data using umap.
\newblock {\em Nature biotechnology}, 37(1):38--44, 2019.

\bibitem{wolf2018scanpy}
F~Alexander Wolf, Philipp Angerer, and Fabian~J Theis.
\newblock Scanpy: large-scale single-cell gene expression data analysis.
\newblock {\em Genome biology}, 19:1--5, 2018.

\bibitem{mccarthy2017scater}
Davis~J McCarthy, Kieran~R Campbell, Aaron~TL Lun, and Quin~F Wills.
\newblock Scater: pre-processing, quality control, normalization and visualization of single-cell rna-seq data in r.
\newblock {\em Bioinformatics}, 33(8):1179--1186, 2017.

\bibitem{xin2016rna}
Yurong Xin, Jinrang Kim, Haruka Okamoto, Min Ni, Yi~Wei, Christina Adler, Andrew~J Murphy, George~D Yancopoulos, Calvin Lin, and Jesper Gromada.
\newblock Rna sequencing of single human islet cells reveals type 2 diabetes genes.
\newblock {\em Cell metabolism}, 24(4):608--615, 2016.

\bibitem{tritschler2022transcriptional}
Sophie Tritschler, Moritz Thomas, Anika B{\"o}ttcher, Barbara Ludwig, Janine Schmid, Undine Schubert, Elisabeth Kemter, Eckhard Wolf, Heiko Lickert, and Fabian~J Theis.
\newblock A transcriptional cross species map of pancreatic islet cells.
\newblock {\em Molecular Metabolism}, 66:101595, 2022.

\bibitem{pyEnsembl}
pyEnsembl.
\newblock \url{https://github.com/openvax/pyensembl?tab=readme-ov-file}.

\bibitem{pyBiomart}
pybiomart.
\newblock \url{https://github.com/jrderuiter/pybiomart}.

\bibitem{satija2015spatial}
Rahul Satija, Jeffrey~A Farrell, David Gennert, Alexander~F Schier, and Aviv Regev.
\newblock Spatial reconstruction of single-cell gene expression data.
\newblock {\em Nature biotechnology}, 33(5):495--502, 2015.

\bibitem{stuart2019comprehensive}
Tim Stuart, Andrew Butler, Paul Hoffman, Christoph Hafemeister, Efthymia Papalexi, William~M Mauck, Yuhan Hao, Marlon Stoeckius, Peter Smibert, and Rahul Satija.
\newblock Comprehensive integration of single-cell data.
\newblock {\em cell}, 177(7):1888--1902, 2019.

\bibitem{he2020single}
Shuai He, Lin-He Wang, Yang Liu, Yi-Qi Li, Hai-Tian Chen, Jing-Hong Xu, Wan Peng, Guo-Wang Lin, Pan-Pan Wei, Bo~Li, et~al.
\newblock Single-cell transcriptome profiling of an adult human cell atlas of 15 major organs.
\newblock {\em Genome biology}, 21:1--34, 2020.

\bibitem{segerstolpe2016single}
{\AA}sa Segerstolpe, Athanasia Palasantza, Pernilla Eliasson, Eva-Marie Andersson, Anne-Christine Andr{\'e}asson, Xiaoyan Sun, Simone Picelli, Alan Sabirsh, Maryam Clausen, Magnus~K Bjursell, et~al.
\newblock Single-cell transcriptome profiling of human pancreatic islets in health and type 2 diabetes.
\newblock {\em Cell metabolism}, 24(4):593--607, 2016.

\bibitem{ma2020chromatin}
Sai Ma, Bing Zhang, Lindsay~M LaFave, Andrew~S Earl, Zachary Chiang, Yan Hu, Jiarui Ding, Alison Brack, Vinay~K Kartha, Tristan Tay, et~al.
\newblock Chromatin potential identified by shared single-cell profiling of rna and chromatin.
\newblock {\em Cell}, 183(4):1103--1116, 2020.

\bibitem{bastidas2019comprehensive}
Aim{\'e}e Bastidas-Ponce, Sophie Tritschler, Leander Dony, Katharina Scheibner, Marta Tarquis-Medina, Ciro Salinno, Silvia Schirge, Ingo Burtscher, Anika B{\"o}ttcher, Fabian~J Theis, et~al.
\newblock Comprehensive single cell mrna profiling reveals a detailed roadmap for pancreatic endocrinogenesis.
\newblock {\em Development}, 146(12):dev173849, 2019.

\bibitem{sharma2018longitudinal}
Ankur Sharma, Elaine~Yiqun Cao, Vibhor Kumar, Xiaoqian Zhang, Hui~Sun Leong, Angeline Mei~Lin Wong, Neeraja Ramakrishnan, Muhammad Hakimullah, Hui Min~Vivian Teo, Fui~Teen Chong, et~al.
\newblock Longitudinal single-cell rna sequencing of patient-derived primary cells reveals drug-induced infidelity in stem cell hierarchy.
\newblock {\em Nature communications}, 9(1):4931, 2018.

\bibitem{aissa2021single}
Alexandre~F Aissa, Abul~BMMK Islam, Majd~M Ariss, Cammille~C Go, Alexandra~E Rader, Ryan~D Conrardy, Alexa~M Gajda, Carlota Rubio-Perez, Klara Valyi-Nagy, Mary Pasquinelli, et~al.
\newblock Single-cell transcriptional changes associated with drug tolerance and response to combination therapies in cancer.
\newblock {\em Nature communications}, 12(1):1628, 2021.

\bibitem{bell2019targeting}
Charles~C Bell, Katie~A Fennell, Yih-Chih Chan, Florian Rambow, Miriam~M Yeung, Dane Vassiliadis, Luis Lara, Paul Yeh, Luciano~G Martelotto, Aljosja Rogiers, et~al.
\newblock Targeting enhancer switching overcomes non-genetic drug resistance in acute myeloid leukaemia.
\newblock {\em Nature communications}, 10(1):2723, 2019.

\bibitem{zheng2017massively}
Grace~XY Zheng, Jessica~M Terry, Phillip Belgrader, Paul Ryvkin, Zachary~W Bent, Ryan Wilson, Solongo~B Ziraldo, Tobias~D Wheeler, Geoff~P McDermott, Junjie Zhu, et~al.
\newblock Massively parallel digital transcriptional profiling of single cells.
\newblock {\em Nature communications}, 8(1):14049, 2017.

\bibitem{the2022tabula}
The Tabula~Sapiens Consortium*, Robert~C Jones, Jim Karkanias, Mark~A Krasnow, Angela~Oliveira Pisco, Stephen~R Quake, Julia Salzman, Nir Yosef, Bryan Bulthaup, Phillip Brown, et~al.
\newblock The tabula sapiens: A multiple-organ, single-cell transcriptomic atlas of humans.
\newblock {\em Science}, 376(6594):eabl4896, 2022.

\bibitem{tabula2020single}
A single-cell transcriptomic atlas characterizes ageing tissues in the mouse.
\newblock {\em Nature}, 583(7817):590--595, 2020.

\bibitem{shazeer2018adafactor}
Noam Shazeer and Mitchell Stern.
\newblock Adafactor: Adaptive learning rates with sublinear memory cost.
\newblock In {\em International Conference on Machine Learning}, pages 4596--4604. PMLR, 2018.

\bibitem{newman2015robust}
Aaron~M Newman, Chih~Long Liu, Michael~R Green, Andrew~J Gentles, Weiguo Feng, Yue Xu, Chuong~D Hoang, Maximilian Diehn, and Ash~A Alizadeh.
\newblock Robust enumeration of cell subsets from tissue expression profiles.
\newblock {\em Nature methods}, 12(5):453--457, 2015.

\bibitem{staahl2016visualization}
Patrik~L St{\aa}hl, Fredrik Salm{\'e}n, Sanja Vickovic, Anna Lundmark, Jos{\'e}~Fern{\'a}ndez Navarro, Jens Magnusson, Stefania Giacomello, Michaela Asp, Jakub~O Westholm, Mikael Huss, et~al.
\newblock Visualization and analysis of gene expression in tissue sections by spatial transcriptomics.
\newblock {\em Science}, 353(6294):78--82, 2016.

\bibitem{levy2016metabolic}
P~Levy and B~Bartosch.
\newblock Metabolic reprogramming: a hallmark of viral oncogenesis.
\newblock {\em Oncogene}, 35(32):4155--4164, 2016.

\bibitem{villani2017single}
Alexandra-Chlo{\'e} Villani, Rahul Satija, Gary Reynolds, Siranush Sarkizova, Karthik Shekhar, James Fletcher, Morgane Griesbeck, Andrew Butler, Shiwei Zheng, Suzan Lazo, et~al.
\newblock Single-cell rna-seq reveals new types of human blood dendritic cells, monocytes, and progenitors.
\newblock {\em Science}, 356(6335):eaah4573, 2017.

\bibitem{wu2020tools}
Yan Wu and Kun Zhang.
\newblock Tools for the analysis of high-dimensional single-cell rna sequencing data.
\newblock {\em Nature Reviews Nephrology}, 16(7):408--421, 2020.

\bibitem{tejada2023causal}
Alejandro Tejada-Lapuerta, Paul Bertin, Stefan Bauer, Hananeh Aliee, Yoshua Bengio, and Fabian~J Theis.
\newblock Causal machine learning for single-cell genomics.
\newblock {\em arXiv preprint arXiv:2310.14935}, 2023.

\bibitem{DBLP:conf/naacl/DevlinCLT19}
Jacob Devlin, Ming{-}Wei Chang, Kenton Lee, and Kristina Toutanova.
\newblock {BERT:} pre-training of deep bidirectional transformers for language understanding.
\newblock In {\em {NAACL-HLT} {(1)}}, pages 4171--4186. Association for Computational Linguistics, 2019.

\bibitem{liu2021machine}
Jiajia Liu, Zhiwei Fan, Weiling Zhao, and Xiaobo Zhou.
\newblock Machine intelligence in single-cell data analysis: advances and new challenges.
\newblock {\em Frontiers in Genetics}, 12:655536, 2021.

\bibitem{oller2021algorithmic}
Sergio Oller-Moreno, Karin Kloiber, Pierre Machart, and Stefan Bonn.
\newblock Algorithmic advances in machine learning for single-cell expression analysis.
\newblock {\em Current Opinion in Systems Biology}, 25:27--33, 2021.

\bibitem{ji2021machine}
Yuge Ji, Mohammad Lotfollahi, F~Alexander Wolf, and Fabian~J Theis.
\newblock Machine learning for perturbational single-cell omics.
\newblock {\em Cell Systems}, 12(6):522--537, 2021.

\bibitem{angerer2017single}
Philipp Angerer, Lukas Simon, Sophie Tritschler, F~Alexander Wolf, David Fischer, and Fabian~J Theis.
\newblock Single cells make big data: new challenges and opportunities in transcriptomics.
\newblock {\em Current opinion in systems biology}, 4:85--91, 2017.

\bibitem{zhao2024langcell}
Suyuan Zhao, Jiahuan Zhang, Yushuai Wu, Yizhen Luo, and Zaiqing Nie.
\newblock Langcell: Language-cell pre-training for cell identity understanding.
\newblock In {\em {ICML}}. OpenReview.net, 2024.

\bibitem{edwards-etal-2022-translation}
Carl Edwards, Tuan Lai, Kevin Ros, Garrett Honke, Kyunghyun Cho, and Heng Ji.
\newblock Translation between molecules and natural language.
\newblock In {\em Proceedings of the 2022 Conference on Empirical Methods in Natural Language Processing}, pages 375--413, Abu Dhabi, United Arab Emirates, December 2022. Association for Computational Linguistics.

\bibitem{fang2023mol}
Yin Fang, Xiaozhuan Liang, Ningyu Zhang, Kangwei Liu, Rui Huang, Zhuo Chen, Xiaohui Fan, and Huajun Chen.
\newblock Mol-instructions: A large-scale biomolecular instruction dataset for large language models.
\newblock {\em ICLR}, 2024.

\bibitem{DBLP:journals/corr/abs-2306-11976}
Zheni Zeng, Bangchen Yin, Shipeng Wang, Jiarui Liu, Cheng Yang, Haishen Yao, Xingzhi Sun, Maosong Sun, Guotong Xie, and Zhiyuan Liu.
\newblock Chatmol: interactive molecular discovery with natural language.
\newblock {\em Bioinformatics}, 40(9):btae534, 2024.

\bibitem{tang2024mollm}
Xiangru Tang, Andrew Tran, Jeffrey Tan, and Mark~B Gerstein.
\newblock Mollm: a unified language model for integrating biomedical text with 2d and 3d molecular representations.
\newblock {\em Bioinformatics}, 40(Supplement\_1):i357--i368, 2024.

\bibitem{singhal2023large}
Karan Singhal, Shekoofeh Azizi, Tao Tu, S~Sara Mahdavi, Jason Wei, Hyung~Won Chung, Nathan Scales, Ajay Tanwani, Heather Cole-Lewis, Stephen Pfohl, et~al.
\newblock Large language models encode clinical knowledge.
\newblock {\em Nature}, 620(7972):172--180, 2023.

\bibitem{boiko2023autonomous}
A.~Boiko Daniil, MacKnight Robert, Kline Ben, and Gomes Gabe.
\newblock Autonomous chemical research with large language models.
\newblock {\em Nature}, pages 570--578, 2023.

\end{thebibliography}

\renewcommand\thefigure{\thesection} 
\renewcommand\thetable{\thesection}

\end{document}